\documentclass[runningheads]{llncs}

\usepackage[mobile]{eccv}

\usepackage{eccvabbrv}
\usepackage{graphicx}
\usepackage{booktabs}
\usepackage{amsmath}
\usepackage{amssymb}
\usepackage{multirow}
\usepackage{xcolor}
\usepackage[accsupp]{axessibility}

\usepackage{hyperref}
\definecolor{linkcolor}{RGB}{255,0,0}
\definecolor{urlcolor}{RGB}{255,105,180}
\definecolor{citecolor}{RGB}{66,168,235}
\hypersetup{colorlinks=true,linkcolor=linkcolor,urlcolor=urlcolor,citecolor=blue}
\usepackage{orcidlink}

\newcommand{\ours}{MotionAtlas}
\newcommand{\ourbench}{\ours-Bench}
\newcommand{\ourdata}{\ours-Data}

\definecolor{upcolor}{RGB}{57,182,74}
\newcommand{\up}[1]{\textcolor{upcolor}{$\uparrow$ #1}}
\newcommand{\down}[1]{\textcolor{red}{$\downarrow$ #1}}

\begin{document}

\title{MotionAtlas: Detailed Region Captioning for Motion-Centric Videos}

\titlerunning{MotionAtlas}

\author{Weisong Liu\inst{1*} \and
Haochen Wang\inst{1*} \and
Kuan Gao\inst{2*} \and
Yuhao Wang\inst{4} \and\\
Yikang Zhou\inst{5} \and
Zhongwei Ren\inst{6} \and
Jacky Mai\inst{3} \and
Anna Wang\inst{3} \and \\
Yanwei Li\inst{2} \and
Jason Li\inst{3\ddagger} \and
Zhaoxiang Zhang\inst{1\dagger}
}

\authorrunning{W.~Liu et al.}

\institute{%
\begin{tabular}{@{}c@{}}
\inst{1}Institute of Automation, Chinese Academy of Sciences \\
\inst{2}Shanghai Jiao Tong University \quad
\inst{3}Nanyang Technological University \\
\inst{4}Peking University \quad
\inst{5}Wuhan University \quad
\inst{6}Beijing Jiaotong University \\
{\small{$^*$Equal contribution \quad
$^\ddagger$Project lead \quad
$^\dagger$Corresponding author}} \\
\vspace{3pt}
Project page: {\small\url{https://kagura-0001.github.io/projects/MotionAtlas}}
\end{tabular}}

\maketitle

\renewcommand{\paragraph}[1]{\vspace{1.25mm}\noindent\textbf{#1}}

\begin{abstract}
We propose \textbf{MotionAtlas}, a system for detailed captioning of \textit{motion-centric} videos, comprising (1) a dedicated human-annotated benchmark, (2) a scalable, high-quality pipeline to construct training samples, and (3) a family of powerful Video-MLLMs. 
Unlike conventional global motion captioning datasets, we focus on region-aware motion captioning: given a video and a spatiotemporal mask, the model generates precise descriptions of motion \textit{within the target region}, thereby alleviating visual clutter and motion entanglement and enabling reliable, quantifiable evaluation.
Concretely, we first build MotionAtlas-Bench, a comprehensive benchmark comprising 2,073 multiple-choice questions, meticulously annotated for a curated set of high-quality, motion-centric videos, to evaluate fine-grained motion understanding of the objects in question.
Second, we design a rigorous and scalable data pipeline that leverages self-bootstrap refinement to suppress fine-grained hallucinations, yielding 159k high-quality motion captioning data.
Third, we design a tailored training data composition strategy, which achieves consistent and substantial performance gains across diverse baseline Video-MLLMs, including Molmo2 and Qwen3-VL.
For instance, MotionAtlas-4B surpasses Qwen3-VL-4B by an average of 5.2 percentage points across general motion benchmarks.
The benchmark, dataset, and code have been released.

\end{abstract}

\section{Introduction}
\label{sec:intro}

\begin{figure}[t]
    \centering\small
    \includegraphics[width=1\linewidth]{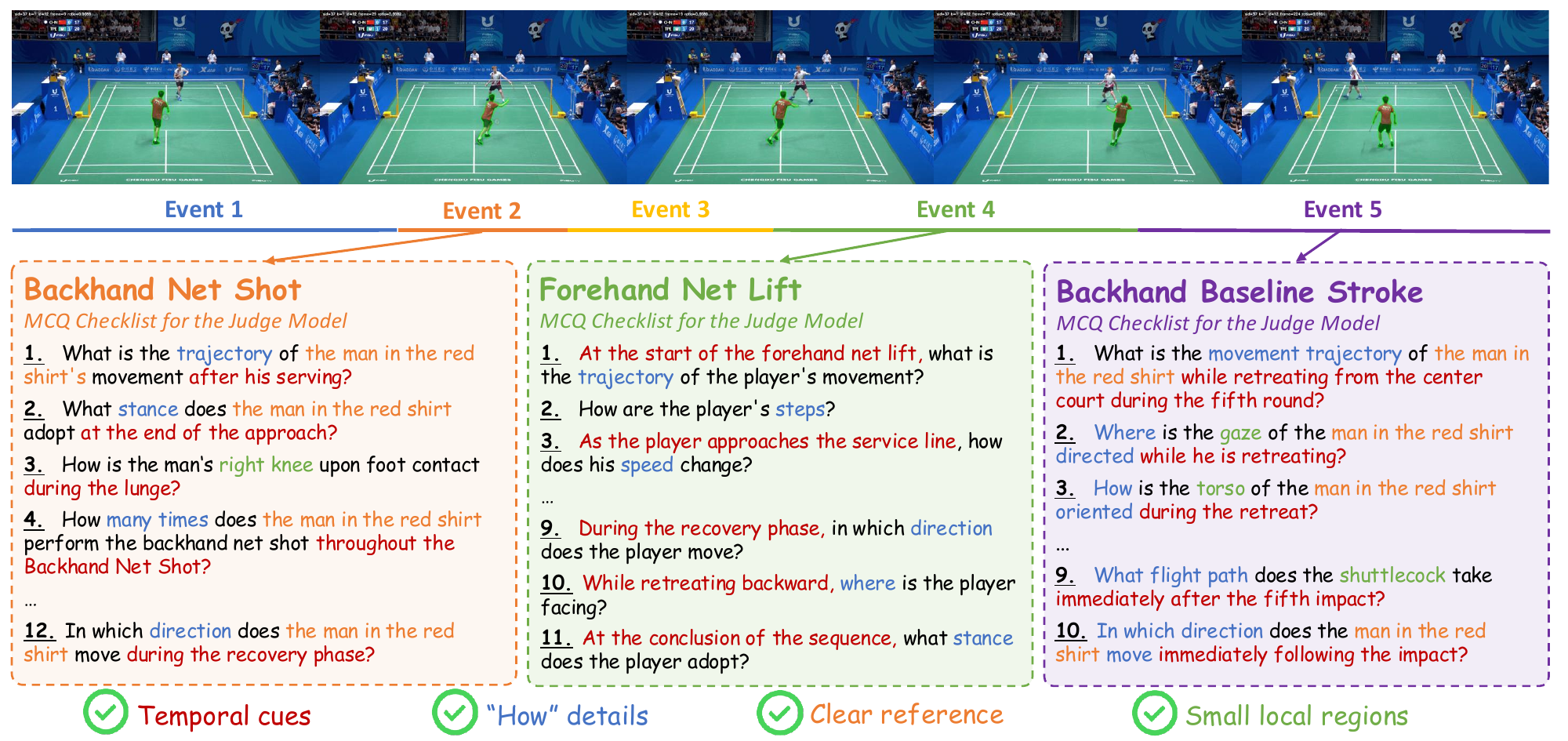}
    \vspace{-17pt}
    \caption{
    Illustration of our MotionAtlas-Bench. 
    Each video is first decomposed into events.
    Then, for each event, the judge model, using candidate captions, answers each multiple-choice question (MCQ) on the checklist.
    The questions emphasize temporal cues, how-level kinematics, clear references, and small local regions, enabling reliable and diagnostic evaluation.
    }
    \label{fig:teaser1}
\end{figure}

\begin{table}[t]
\centering\small
\caption{Comparison of related datasets.
\textbf{Part I} compares MotionAtlas-Bench against existing evaluation benchmarks, emphasizing dense QA coverage and motion attribute details. 
``$r_t$'' denotes relative object size (mean / median).
``Mot. Score'' represents the magnitude of the Farneback optical flow. 
``MCQ'' means multiple-choice question.
\textit{Our MotionAtlas-Bench is comprehensive and challenging.}
\textbf{Part II} compares MotionAtlas-Data with existing training datasets in terms of caption complexity and verb density.
``Qua. Con.'' indicates quality control, and our data incorporates a self-bootstrap refinement strategy introduced in \S\ref{sec:scbs}.
\textit{Our MotionAtlas-Data is detailed and faithful.}
}
\label{tab:bench_compare}
\vspace{-10pt}

\resizebox{\linewidth}{!}{
\begin{tabular}{l c c c c c c}
\toprule
\multicolumn{7}{l}{\textit{\textbf{Part I. Evaluation Benchmarks}}} \\
\midrule
\textbf{Benchmark} & \textbf{Ref. Obj.} & \textbf{Mot. Score} & \textbf{Anno. Type} & \textbf{QA / Vid.} & \textbf{Obj. Size ($r_t$)} & \textbf{Multi-Aspect Attr.} \\
\midrule
MotionBench~\cite{hong2025motionbench} & $\times$ & 3.6 & QA & 1.5 & Global & $\times$ \\
FAVOR-Bench~\cite{tu2025favorbench} & $\times$ & 4.6 & Caption / MCQ & 4.6 & Global & $\times$ \\
DREAM-1K~\cite{wang2024tarsierrecipestrainingevaluating} & $\times$ & 4.5 & Caption & -- & Global & $\times$ \\
VideoRefer-D & $\checkmark$ & 4.2 & Caption & -- & 0.17 / 0.14 & $\times$ \\
\textbf{MotionAtlas-Bench} & $\checkmark$ & \textbf{5.4} & \textbf{MCQ} & \textbf{19.4} & \textbf{0.14 / 0.10} & $\checkmark$ \\
\end{tabular}
}
\resizebox{\linewidth}{!}{
\begin{tabular}{l c c c c c c c c}
\midrule
\multicolumn{9}{l}{\textit{\textbf{Part II. Training Datasets}}} \\
\midrule
\textbf{Dataset} & \textbf{Ref. Obj.} & \textbf{Anno. Method} & \textbf{Anno. Type} & \textbf{Videos} & \textbf{Texts} & \textbf{Verb / Vid.} & \textbf{Length} & \textbf{Qua. Con.} \\
\midrule
MotionSight~\cite{du2025motionsight} & $\times$ & Auto & QA & 40K & 87K & 12 & 70 & -- \\
VideoRefer (\textit{Detail})~\cite{yuan2025videorefer} & $\checkmark$ & Auto & Caption & 125K & 125K & 11 & 65 & -- \\
TarsierRecap~\cite{wang2024tarsierrecipestrainingevaluating} & $\times$ & Auto & Caption & 585K & 585K & 13 & 74 & -- \\
MotionBench (Train)~\cite{hong2025motionbench} & $\times$ & Human & Caption & 5K & 5K & 17 & 121 & Human \\
\textbf{MotionAtlas-Data} & $\checkmark$ & \textbf{Auto} & \textbf{Caption} & \textbf{82K} & \textbf{159K} & \textbf{23} & \textbf{212} & \textbf{Self-Bootstrap} \\
\bottomrule
\end{tabular}
}
\vspace{-10pt}
\end{table}

\begin{figure}[t]
    \centering\small
    \includegraphics[width=1\linewidth]{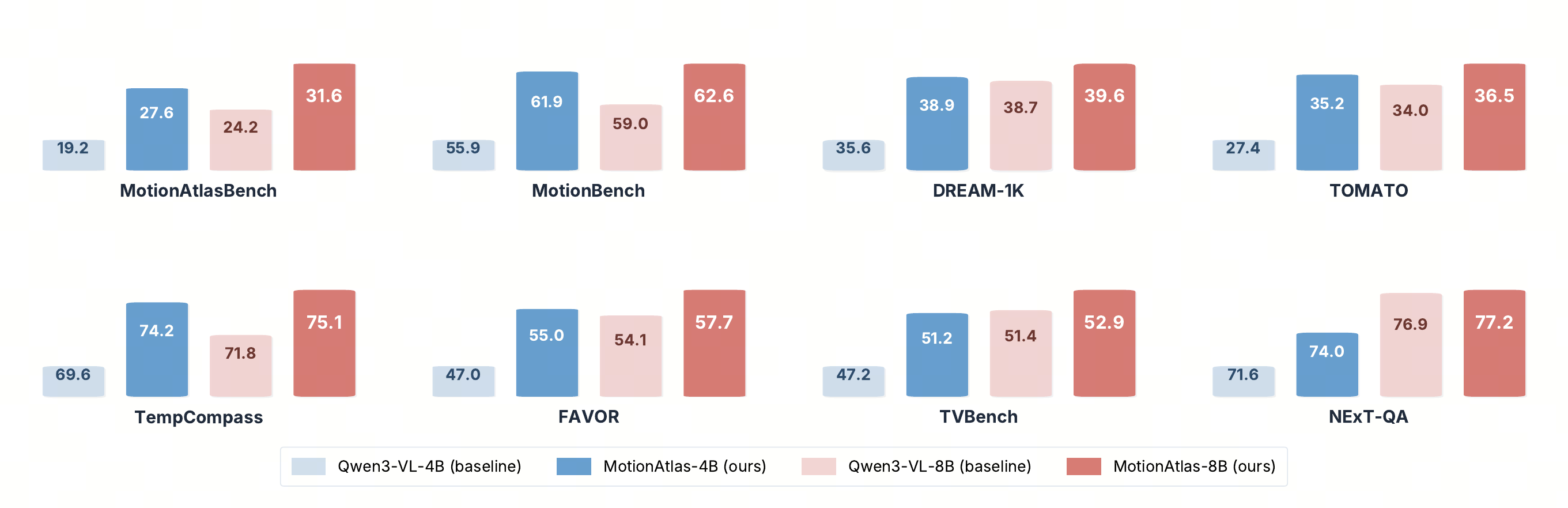}
    \vspace{-20pt}
    \caption{
    Quantitative comparison between our MotionAtlas and baselines (Qwen3-VL).
    Our MotionAtlas brings significant improvements over baselines \textit{consistently}.
    }
    \label{fig:exp}
    \vspace{-10pt}
\end{figure}

Driven by high-quality video-text datasets and large-scale model development, Video Multimodal Large Language Models (Video-MLLMs)~\cite{wang2025internvl3, clark2026molmo2, gemmateam2025gemma3technicalreport, bai2025qwen3vltechnicalreport} have achieved significant progress on general video understanding tasks. 
However, most existing models still struggle with fine-grained motion dynamics in real-world videos, particularly when precise spatiotemporal reasoning over local regions is required~\cite{zhang2025videollama3frontiermultimodal,wang2025internvideo25empoweringvideomllms}.
For instance, the professional badminton scenario in Figure~\ref{fig:teaser1} requires models to answer diagnostic questions that demand fine-grained motion perception, \textit{e.g.}, trajectory tracking, kinematic state assessment, and temporal anchoring.
Such tasks remain difficult for current Video-MLLMs, which tend to prioritize high-level semantics over fine-grained, context-aware motion dynamics.

Most prior works focus on global video motion captioning, which aims to describe the motion of the entire scene or prominent objects~\cite{yuan2025tarsier2,tu2025favorbench, wang2024tarsierrecipestrainingevaluating,chen2024sharegptvideo}. 
While this paradigm has been widely explored, its core limitation lies in its \textit{inherent evaluation intractability}: reliably assessing the quality of generated global motion captions is extremely challenging, which makes it hard to verify their completeness (recall) and faithfulness (anti-hallucination)~\cite{wang2024tarsierrecipestrainingevaluating,chai2025auroracap,hong2025motionbench}, leading to unreliable annotation pipelines that cannot scale.

To address this, we shift the focus to \textit{region} captioning~\cite{yuan2025videorefer,yuan2025pixelreferunifiedframeworkspatiotemporal,chen2023shikraunleashingmultimodalllms,yuan2024osprey,zhang2024gptroi,guo2024regiongpt} for motion-centric videos: given a video and a spatiotemporal mask, the model generates a precise description of the motion \textit{within the target region only}.
Confining the description to a spatiotemporal mask removes the visual clutter and motion entanglement inherent to global captioning, allowing annotators to focus exclusively on local motion.
This yields finer-grained, more complete, and more consistent annotations, which in turn enable the reliable, quantifiable, and scalable evaluation that is otherwise infeasible for global motion description.
To this end, we propose \textbf{MotionAtlas}, comprising a human-annotated evaluation benchmark, a carefully designed training corpus, and a family of powerful Video-MLLMs.

As the foundation, we first construct MotionAtlas-Bench, tailored for region-level motion captioning, to systematically validate the quality of generated motion descriptions and to ablate the effectiveness of candidate data pipelines.
Because our annotations are equipped with fine-grained identity references and temporal anchors, we can reliably localize missing motion details and diagnose fine-grained hallucinations in the generated descriptions~\cite{wang2024tarsierrecipestrainingevaluating,ding2023mevis,hong2025motionbench,xu2022finediving}. 
As summarized in Table~\ref{tab:bench_compare} (Part~I), our benchmark is both:
\begin{itemize}
    \item \textit{Comprehensive.} It provides far denser supervision than prior benchmarks, with 19.4 questions per video, and is the only benchmark to annotate detailed multi-aspect motion attributes.
    \item \textit{Challenging.} It targets smaller objects that are harder to localize and track, and exhibits the highest motion complexity among all compared benchmarks (Table~\ref{tab:bench_compare}).
\end{itemize}

Next, we design a rigorous and scalable training data pipeline that operates in three stages.
First, temporal segmentation produces detailed motion caption proposals for each segment. 
Second, a self-bootstrap refinement mechanism contrasts multiple rollout captions within the same temporal window to identify high-risk claims~\cite{manakul2023selfcheckgpt}, and then exploits the VLM's stronger understanding than generation~\cite{yang2025captionqa,manakul2023selfcheckgpt,wang2022self} to filter out implausible details and hallucinations. 
Third, multi-source summarization guided by global captions as temporal cues removes inconsistencies and redundancies, yielding coherent motion descriptions~\cite{hong2025motionbench,xu2022finediving,yuan2025tarsier2,cho2025perceptionlm,sun2022dance,dave2020tao,lin2025perceive}. 
As shown in Table~\ref{tab:bench_compare} (Part~II), this pipeline produces richer, longer motion annotations while substantially reducing fine-grained hallucinations compared with existing datasets.

Finally, to fully leverage this data, we propose a tailored training recipe that strengthens the model's ability to capture region-specific motion dynamics.
By integrating spatiotemporal region cues into the training mixture, we obtain a family of Video-MLLMs that excel at fine-grained video understanding, not limited to region motion captioning.

Empirically, as demonstrated in Figure~\ref{fig:exp}, MotionAtlas brings \textit{consistent} and significant improvements over various baselines~\cite{clark2026molmo2, bai2025qwen3vltechnicalreport} across eight representative motion-related and general video understanding benchmarks.

\section{\ourbench{}}
\label{sec:bench}

\ourbench{} operationalizes region-level motion evaluation through an \emph{event--motion--fact} annotation hierarchy: each video is split into events, each event is described in dense motion detail, and each detail is distilled into an atomic, independently verifiable fact. The resulting facts form a multiple-choice-question (MCQ) checklist for judging a caption, which turns the otherwise intractable problem of scoring free-form motion descriptions into dense, diagnostic verification.

\subsection{Data Curation}
\label{sec:data_curation}

We draw candidate clips from four motion-rich sources~\cite{wang2024tarsierrecipestrainingevaluating,ding2023mevis,hong2025motionbench,xu2022finediving} and retain only entities whose motion is genuinely difficult to describe. A VLM scores each candidate and keeps those exhibiting salient multi-atomic motion or small object/background motion, precisely the cases in which global captioning is least reliable. We further include task-oriented clips covering complex kinematics (\eg, somersaults, footwork sequences) and rapid, multi-step motion, so that the benchmark concentrates on challenging dynamics rather than simple single-action clips. Where source masks are missing, annotators supplement them with SAM2~\cite{ravi2024sam} to guarantee per-frame spatial grounding for every referred entity.

\subsection{Evaluation Protocol}
\label{sec:eval_protocol}

\paragraph{Checklist-Based Judgement.}
To holistically evaluate fine-grained motion captioning, we perform dense evaluation against granular factual questions (as exemplified in Figure~\ref{fig:teaser1}).
Given a model-generated caption $C$ for video $V$, a judge model is presented with \textit{each} MCQ $\hat{Q}_k$ alongside $C$, and selects the most appropriate option.
%
%
Based on the judge's selections across all MCQs, we report three complementary metrics:
\begin{itemize}

  \item \textbf{Accuracy} $= N_{\text{correct}} / N_{\text{total}}$: the overall proportion of correctly described motions across all test cases.

  \item \textbf{Recall} $= (N_{\text{correct}} + N_{\text{error}}) / N_{\text{total}}$: the tendency to describe the target motion rather than respond ``not mentioned''.

  \item \textbf{Precision} $= N_{\text{correct}} / (N_{\text{correct}} + N_{\text{error}})$: the accuracy of the descriptions when the motion is explicitly mentioned.

\end{itemize}

We design two increasingly difficult grounding settings to disentangle different aspects of model capability.

\paragraph{Single-Frame Grounding.}
The model receives the full video and only the target entity's mask at its first visible frame, so it must autonomously track the entity throughout the clip, jointly testing spatial tracking and motion understanding.

\paragraph{Full-Sequence Grounding.}
The model receives the full video with exact per-frame masks overlaid on every frame. This continuous localization removes tracking ambiguity and background distractors, strictly measuring the model's intrinsic motion captioning capacity.

\subsection{Annotation Pipeline}
\label{sec:anno_pipeline}

%
%
\paragraph{Overview.}
Fine-grained motion is continuous, layered, and partly subjective, giving rise to three recurring failure modes of naive annotation: ambiguous event boundaries, concurrent body-part motions, and hard-to-verify attributes (\eg, speed, force). Our pipeline addresses them in turn, pairing a VLM for coverage with a human for correction at every stage:
\begin{equation}\label{eq:bench_anno}
     V,\{m_t\}
  \;\xrightarrow{\text{Segment}}\; \{\hat{E}_k\}
  \;\xrightarrow{\text{Caption}}\; \{D_k\}
  \;\xrightarrow{\text{MCQ Extraction}}\; \{\hat{Q}_k\},
\end{equation}
where $V$ is the video, $m_t$ the target mask at frame $t$, and $k$ indexes events; $\hat{E}_k$ is the human-refined event, $D_k$ its dense motion description, and $\hat{Q}_k$ the verified MCQ entering the checklist. Hats mark artifacts finalized by human review.

\paragraph{Event Annotation.}
%
%
Segmenting a long clip into events lets both the VLM and the annotator reason over a short window at high temporal resolution rather than over the entire video at once. We enforce a single principle, \textit{one event corresponds to one core motion}, which directly resolves the boundary ambiguity noted above. Gemini 3 Pro proposes events with frame intervals; annotators then merge over-fragmented events, correct inaccurate boundaries, and remove hallucinated motion (\eg, camera motion attributed to the subject).
%

\paragraph{Detailed Fact Proposal and Annotation.}
Events carry only coarse semantics, so we densify each one along a motion ontology of six aspects: \textsc{Kinematics} (direction, trajectory, speed change, rotation), \textsc{Parts} (independent body-part motions, \eg, ``left fist jabs forward while right hand guards the chin''), \textsc{Spatial} (position relative to environmental landmarks), \textsc{Interaction} (contact, grasping, releasing), \textsc{State} (posture and occlusion changes), and \textsc{Camera} (camera motion).
%

For each event, we feed the VLM a \emph{merged focal crop}, the union of the entity's bounding boxes across the whole event rather than per-frame crops, so the crop region stays fixed even under large displacement: this removes spatial jitter and raises the effective resolution on the target while suppressing background distractors.
%
The VLM drafts a dense description, and annotators then supplement the details that VLMs systematically omit, chiefly part-level actions and precise directions.
%

\paragraph{Multiple-Choice Questions Construction.}
%
To make scoring objective, we decompose each dense description into atomic facts, each isolating a single independently verifiable attribute. Gemini 3 Pro then converts every fact into an MCQ whose distractors are alternative values \emph{within the same aspect} (\eg, a different rotation direction), requiring the judge to discriminate fine motion details rather than coarse topics; the option structure and metrics follow \S\ref{sec:eval_protocol}.



\paragraph{Quality Control.}
We carry out two-tier quality control: (1) \textit{Blind filtering}: an LLM answers questions without video context, removing low-discriminability questions that can be guessed by commonsense.
(2) \textit{Human fact verification}: annotators verify each correct option's accuracy, correct hallucinated attributes, ensure at least one hard-case confusion option, and assign missing options to unverifiable facts.
Each MCQ is reviewed by two annotators: factual-validity disagreements are removed or marked as unverifiable, while wording or option-design disagreements are resolved by adjudication.
%

\subsection{Benchmark Statistics}
\label{sec:bench_stats}

\begin{figure*}[!t]
    \centering
    \begin{minipage}[c]{0.45\textwidth}
        \centering
        \includegraphics[width=\linewidth]{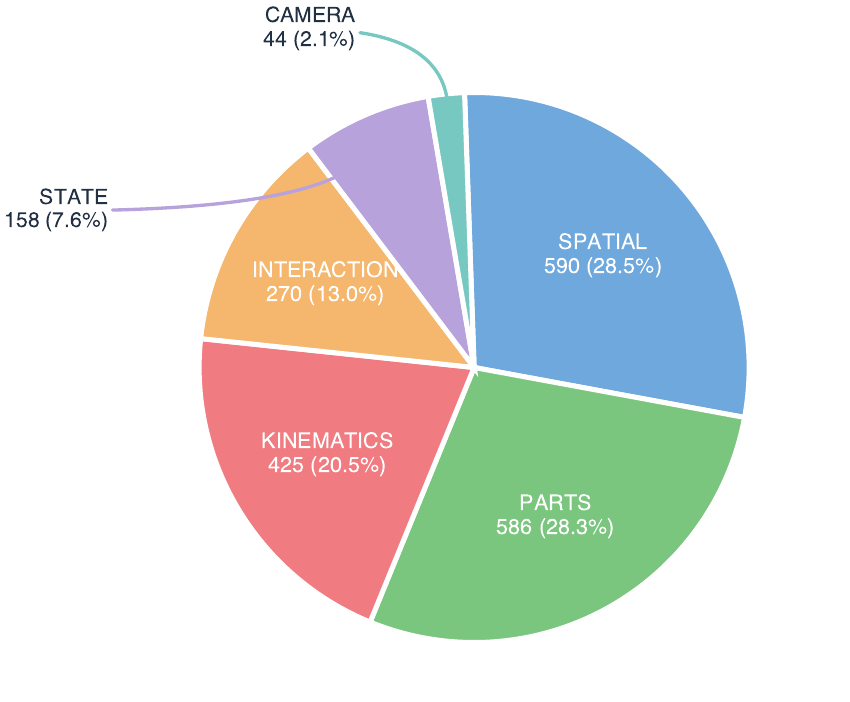}
        \vspace{-24pt}
        \caption{Data distribution of our proposed MotionAtlas-Bench aspects.}
        \label{fig:data_source}
    \end{minipage}
    \begin{minipage}[c]{0.53\textwidth}
        \centering\small
        \makeatletter\def\@captype{table}\makeatother 
        
        \caption{
        \ourdata{} sources.
        Details of our filtering strategy are introduced in \S\ref{sec:data_collection}.
        %
        }
        \label{tab:data_sources}
        \vspace{-10pt}
        \setlength{\tabcolsep}{8pt} 
        \begin{tabular}{lrr}
        \toprule
        Data Source & Videos & Samples \\
        \midrule
        MeVIS~\cite{ding2023mevis}      & 1.6K  & 8K   \\
        SAV~\cite{ravi2024sam}        & 39K   & 82K  \\
        TAO~\cite{dave2020tao}        & 500   & 1.4K \\
        DanceTrack~\cite{sun2022dance} & 37    & 210  \\
        ViCaS~\cite{athar2025vicas}      & 20K   & 65K  \\
        VastTrack~\cite{peng2024vasttrack}  & 46K   & 46K  \\
        GOT~\cite{huang2019got}        & 10K   & 18K  \\
        \midrule
        Total (Raw) & 126K & 220K \\
        \textbf{Total} (Filtered) & \textbf{82K} & \textbf{159K}\\
        \bottomrule
        \end{tabular}
    \end{minipage}
    \vspace{-10pt}
\end{figure*}

After quality control, MotionAtlas-Bench contains 107 videos and 2,073 MCQs. The aspect distribution (Figure~\ref{fig:data_source}) is dominated by \textsc{Spatial} (29.7\%), \textsc{Parts} (27.7\%), and \textsc{Kinematics} (19.8\%), reflecting our emphasis on fine-grained body articulations and spatial dynamics over coarse action labels. Most videos (57.0\%) span 30--60s and 39.3\% exceed one minute, requiring models to continuously track the target and its fine-grained motion throughout the clip. Each video averages 19.4 MCQs (max 56), ensuring dense coverage of concurrent motion layers.
Target entities are small, with a mean rate of 14\% as shown in Table~\ref{tab:bench_compare}, requiring precise spatial grounding rather than reliance on global scene cues.

\section{Data Pipeline and Training Recipe for MotionAtlas}
\label{sec:method}

\subsection{Data Collection \& Filtering}
\label{sec:data_collection}

We collect videos with per-frame tracking annotations (mask or bounding box) from 7 diverse open-source datasets, totaling 126K videos and 220K region samples.
Each sample is a (video, entity) pair, and a single video may contain multiple referred entities.
Before annotation, we apply three-tier filtering to the 220K raw samples to ensure data quality.

\paragraph{Basic Quality Filtering.}
We first remove videos shorter than 1 second or longer than 1 minute (clips that are too short lack sufficient motion, while overly long clips exceed the VLM's effective context window), as well as samples whose spatial resolution falls below a minimum threshold.

\paragraph{Motion Score Filter.}
%
Unlike methods based on whole-video global motion scoring, we design an \textit{entity-centric motion score} that measures only the referred entity's own motion intensity while suppressing camera motion and common-mode background motion. Specifically, we compute the optical flow intensity $c_t$ within the entity mask and subtract the flow intensity $\beta_t$ of the surrounding ring and background regions, yielding a residual motion score:
\begin{equation}
  f_t = \max(0,\; c_t - \beta_t) + \alpha \cdot c_t
  \label{eq:motion_score}
\end{equation}
The $\alpha$ term preserves a small absolute motion signal to prevent over-suppression from eliminating genuine motion of small targets. We set $\alpha=0.3$ in all experiments. The final entity motion score jointly considers motion intensity ($90^{\text{th}}$ percentile) and motion persistence (fraction of frames $p$ exceeding a fixed threshold), suppressing sporadic motion spikes.

After three-tier filtering, 159K samples are retained for subsequent annotation.
Detailed sources are illustrated in Table~\ref{tab:data_sources}.

\subsection{Dataset Annotation Pipeline}
\label{sec:scbs}

\noindent\textbf{Overview.}
The benchmark pipeline in~\S\ref{sec:anno_pipeline} achieves high annotation quality through AI-Human collaboration, but its human refinement (event refinement) and human fact verification (MCQ verification) steps limit scalability.
To expand the annotation scale from the benchmark's 107 videos to 159K samples, we need to replace these manual steps with automated mechanisms while maintaining comparable annotation quality.

The core challenge is that, once human review is removed, single-pass VLM captioning introduces systematic hallucinations, particularly at the temporal boundaries between events.
To address this, we propose a four-stage pipeline, \textit{segment--caption--bootstrap--summarize}, which inherits the event segmentation and spatial crop design from~\S\ref{sec:anno_pipeline} and introduces \textit{dual-rollout differential verification} to replace human review and \textit{multi-source narrative synthesis} to eliminate temporal incoherence between events. The entire pipeline uses Qwen3-VL-235B as the unified VLM backbone:
\begin{equation}
  V, \{m_t\} \xrightarrow{\text{Segment}} \{E_k\} \xrightarrow{\text{Caption}} \{c^{(1)}_k, c^{(2)}_k\} \xrightarrow{\text{Bootstrap}} \{\hat{c}_k\} \xrightarrow{\text{Summary}} N
  \label{eq:scbs}
\end{equation}
where $V$ is the input video, $\{m_t\}$ denotes the per-frame entity masks, $E_k$ is the $k$-th segmented motion event, $\hat{c}_k$ is the refined caption for event $E_k$, and $N$ is the final structured motion narrative.

\paragraph{Event Segmentation.}
Following~\S\ref{sec:anno_pipeline}, we segment videos by motion semantics to achieve temporal zoom-in, obtaining event sequences $\{E_k = (s_k, e_k, d_k^{\text{seg}})\}$, where $s_k$ and $e_k$ are the start and end timestamps of event $E_k$, and $d_k^{\text{seg}}$ is a coarse event descriptor generated from sparsely sampled global frames. Here, \textit{no human refinement} is performed.
%
%

\paragraph{Local \& Global Captioning.}
We query the VLM backbone for both local and global captioning.
(1) At the local level, for each event $E_k$ we construct a merged focal crop $V_k^{\text{crop}}$ from the union of bounding boxes across all masks within the event's frame interval, and the VLM generates a local description $c_k$ rich in motion details.
(2) At the global level, the VLM generates a full-video caption $c_{\text{full}}$ that provides scene context and a global timeline.
Local captions offer higher spatiotemporal resolution on motion details, while the global caption provides a reliable overall temporal framework.

\paragraph{Self-Bootstrap Refinement.}
This is the core mechanism that replaces human review in~\S\ref {sec:anno_pipeline}. 
Its design is based on the following observation: \textit{divergences between two independent rollouts are more likely to originate from hallucinations, while agreement points are more likely to reflect genuine visual evidence}. 
%
%
%
This procedure includes four phases:
\begin{enumerate}
  \item Dual Rollout: independently generate two descriptions $\{c_k^{(1)}, c_k^{(2)}\}$ for the same visual input $V_k^{\text{crop}}$.
  \item Differential Extraction: an LLM extracts the set of conflicting claims, which carry the highest uncertainty and are thus the most informative to verify.
  \item Visual Grounded Judgment: for each conflicting claim, the two assertions plus a distractor option are randomly shuffled and presented to the VLM for 3-way blind adjudication.
  \item Caption Correction: verified judgments are applied to the original description to produce the refined result $\hat{c}_k$.
\end{enumerate}

\paragraph{Multi-Source Narrative Synthesis.}
Finally, this stage integrates local captions $\{\hat{c}_k\}$ with the full-video caption $c_{\text{full}}$, removing speculative descriptions at event boundaries and redundant content between adjacent events, while using $c_{\text{full}}$ as a global timeline reference to establish temporal coherence.
%

\subsection{Data Statistics}
\label{sec:data_stats}

\paragraph{Comparison with Existing Training Datasets.}
As shown in Table~\ref{tab:bench_compare} (Part II), MotionAtlas-Data substantially surpasses existing video training datasets in motion density, with an average of 23 action verbs and 212 words per sample.
Crucially, whereas most large-scale datasets provide global descriptions, our annotations are strictly region-level, ensuring that these rich motion details are explicitly grounded to specific target entities.

\subsection{Training Recipe}
\label{sec:recipe}

To validate the effectiveness of \ourdata{} for both general motion and referring (region-level) motion understanding, we mix the following four categories of data during a unified SFT stage, as demonstrated in Table~\ref{tab:data_sft}.

\begin{table*}[t]
\centering\small
\begin{minipage}[t]{0.46\textwidth}
\centering
\makeatletter\def\@captype{table}\makeatother
\caption{Training data mixture for the unified motion SFT stage, spanning general and region-level sources.}
\label{tab:data_sft}
\label{tab:recipe}
\vspace{2pt}
\setlength{\tabcolsep}{6pt}
\resizebox{\linewidth}{!}{%
\begin{tabular}{lll}
\toprule
Category & Source & Scale \\
\midrule
General Motion Caption & Tarsier2-Recap~\cite{yuan2025tarsier2} & 417K \\
General Image/Text QA & LLaVA-OneVision-1.5~\cite{an2025llava} & 320K \\
General Motion QA & MotionSight~\cite{du2025motionsight} & 130K \\
\textbf{Region Motion Caption} & \textbf{MotionAtlas} & \textbf{159K} \\
\bottomrule
\end{tabular}}
\end{minipage}\hfill
\begin{minipage}[t]{0.52\textwidth}
\centering
\makeatletter\def\@captype{table}\makeatother
\caption{Transfer ablation of the SFT data composition on general motion benchmarks (Qwen3-VL-4B).}
\label{tab:transfer_ablation}
\vspace{2pt}
\setlength{\tabcolsep}{4pt}
\resizebox{\linewidth}{!}{%
\begin{tabular}{lccc}
\toprule
MotionAtlas SFT variant & MotionBench & TOMATO & FAVOR \\
\midrule
Base                        & 55.9 & 27.4 & 47.0 \\
+ General caption           & 60.5 \up{4.6} & 28.4 \up{1.0} & 52.2 \up{5.2} \\
+ Region detail, text ref.  & 61.7 \up{5.8} & 31.9 \up{4.5} & 55.7 \up{8.7} \\
+ Region detail, visual cue & \textbf{61.9} \up{6.0} & \textbf{35.2} \up{7.8} & 55.0 \up{8.0} \\
\bottomrule
\end{tabular}}
\end{minipage}
\end{table*}

The mixing strategy pursues three goals: (1) injecting entity-centric fine-grained motion captioning through \ourdata{}; (2) preserving general motion understanding through Tarsier2-Recap~\cite{yuan2025tarsier2} and MotionSight~\cite{du2025motionsight}; and (3) preventing catastrophic forgetting through general QA data from LLaVA-OneVision-1.5~\cite{an2025llava}.
In the unified SFT stage, we concatenate the four sources and sample examples proportionally to dataset size for one epoch.
We analyze the transfer contribution of each data component (Table~\ref{tab:transfer_ablation}) in \S\ref{sec:transfer_ablation}.
%

\section{Experiments}
\label{sec:exp}
%

\begin{table*}[!t]
\centering\small
\caption{Main results on our \textbf{MotionAtlas-Bench}. 
We report results in terms of \textbf{Accuracy (acc.)}.
``Single-Frame'' denotes Single-Frame Grounding and ``Full-Seq.'' denotes Full-Sequence Grounding.
Based on Qwen3-VL-8B~\cite{bai2025qwen3vltechnicalreport}, MotionAtlas-Data even surpasses Qwen3-VL-235B~\cite{bai2025qwen3vltechnicalreport}, demonstrating the effectiveness of our data.
}
\label{tab:bench_results}
\vspace{-10pt}
\setlength{\tabcolsep}{1.5pt} 
\resizebox{\textwidth}{!}{
\begin{tabular}{l c@{\hspace{1em}}cccccc c@{\hspace{1em}}cccccc}
\toprule
\multirow{2}{*}{\textbf{Model}} & \multicolumn{7}{c}{\textbf{Single-Frame Grounding (acc.)}} & \multicolumn{7}{c}{\textbf{Full-Sequence Grounding (acc.)}} \\
\cmidrule(lr){2-8} \cmidrule(lr){9-15}
& \textbf{Overall} & \textbf{Spatial} & \textbf{Parts} & \textbf{Kin.} & \textbf{Inter.} & \textbf{State} & \textbf{Cam.} & \textbf{Overall} & \textbf{Spatial} & \textbf{Parts} & \textbf{Kin.} & \textbf{Inter.} & \textbf{State} & \textbf{Cam.} \\
\midrule
Gemini 3 Pro & 36.4 & \textbf{36.6} & \textbf{34.7} & 32.0 & \textbf{43.2} & 46.0 & 24.0 & 36.5 & \textbf{36.8} & 33.5 & \textbf{38.1} & 38.1 & 44.0 & 22.0 \\
GPT-5.2~\cite{openai_gpt5_2} & \textbf{36.9} & 36.5 & 34.0 & \textbf{34.2} & 42.1 & \textbf{51.3} & 24.0 & \textbf{37.6} & 36.0 & \textbf{38.8} & 36.6 & \textbf{38.5} & 46.7 & 22.0 \\
Gemini 2.5 Pro~\cite{comanici2025gemini25pushingfrontier} & 29.8 & 29.8 & 28.9 & 27.5 & 29.3 & 43.3 & 20.0 & 31.9 & 33.4 & 30.5 & 25.5 & 34.1 & 47.3 & 26.0 \\
Gemini 2.5 Flash~\cite{comanici2025gemini25pushingfrontier} & 28.9 & 29.0 & 27.1 & 25.3 & 33.0 & 40.0 & 24.0 & 35.4 & 34.7 & 36.3 & 32.3 & 35.5 & 47.3 & 22.0 \\
Qwen3-VL-235B~\cite{bai2025qwen3vltechnicalreport} & 30.5 & 31.8 & 27.8 & 28.9 & 31.5 & 38.7 & 30.0 & 33.7 & 35.0 & 33.2 & 31.1 & 34.4 & 38.7 & 28.0 \\
Qwen3-VL-32B~\cite{bai2025qwen3vltechnicalreport} & 29.7 & 30.3 & 27.5 & 26.0 & 31.5 & 41.6 & \textbf{32.0} & 35.7 & 32.8 & 33.9 & 34.4 & 38.4 & \textbf{53.4} & \textbf{36.0} \\
\midrule

Molmo2-4B~\cite{clark2026molmo2} & 13.7  & 13.6 & 12.9 & 13.7 & 14.5 & 16.1 & 13.0 & 15.6 & 15.6 & 15.0 & 15.3 & 16.1 & 18.1 & 14.6 \\

\ \ + MotionAtlas-Data & 21.6\rlap{\ \up{7.9}}& 21.6 & 20.9 & 21.2 & 22.7 & 24.4 & 20.3 & 25.5\rlap{\ \up{9.9}} & 25.4 & 24.8 & 25.2 & 26.4 & 28.3 & 24.5 \\

Molmo2-8B~\cite{clark2026molmo2} & 19.2 & 19.1 & 18.3 & 19.3 & 20.0 & 21.6 & 18.4 & 21.5 & 21.5 & 21.0 & 21.4 & 21.9 & 23.9 & 20.4 \\

\ \ + MotionAtlas-Data & 24.4\rlap{\ \up{5.2}}& 24.2 & 23.4 & 24.3 & 25.4 & 27.4 & 23.0 & 27.5\rlap{\ \up{6.0}}& 27.4 & 26.7 & 27.3 & 28.4 & 30.4 & 26.7 \\

\midrule


Qwen3-VL-4B~\cite{bai2025qwen3vltechnicalreport} & 19.3 & 19.5 & 20.0 & 14.1 & 20.2 & 28.7 & 16.3 & 21.7 & 21.9 & 22.4 & 16.5 & 22.6 & 31.1 & 18.7 \\

\ \ + MotionAtlas-Data & 27.7\rlap{\ \up{8.4}} & 25.9 & 27.9 & 26.9 & 28.3 & 35.3 & 28.2 & 30.1\rlap{\ \up{8.4}}& 28.3 & 30.3 & 29.3 & 30.7 & 37.7 & 30.6 \\

Qwen3-VL-8B~\cite{bai2025qwen3vltechnicalreport} & 24.3 & 24.5 & 23.9 & 20.3 & 24.8 & 37.6 & 18.0 & 26.7 & 27.2 & 24.6 & 26.7 & 26.4 & 34.7 & 22.0 \\

\ \ + MotionAtlas-Data & 31.6\rlap{\ \up{7.3}}& 31.1 & 31.2 & 30.6 & 29.0 & 42.8 & 34.5 & 34.1\rlap{\ \up{7.4}}& 33.5 & 33.6 & 33.0 & 31.4 & 45.2 & 36.9 \\
\bottomrule
\end{tabular}}
\end{table*}


\subsection{Experimental Setup}
\label{sec:exp_setup}

\paragraph{Implementation Details.}
We adopt Qwen3-VL (4B, 8B)~\cite{bai2025qwen3vltechnicalreport} and Molmo2 (4B, 8B)~\cite{clark2026molmo2} as our base models. 
For all models, we perform full parameter fine-tuning for 1 epoch using the data mixture detailed in \S\ref{sec:recipe}. 
During training, we uniformly sample 16 frames per video and limit the maximum sequence length to 16,384 tokens. 
The models are optimized using AdamW with a peak learning rate of $1 \times 10^{-5}$ and a cosine learning rate schedule with a 3\% linear warmup. 

\paragraph{Evaluation Benchmarks.}
To systematically assess fine-grained motion understanding and general video reasoning, we evaluate on 8 representative benchmarks: MotionAtlas-Bench, MotionBench~\cite{hong2025motionbench}, DREAM-1K~\cite{wang2024tarsierrecipestrainingevaluating}, TOMATO~\cite{shangguan2025tomato}, NExT-QA~\cite{xiao2021next}, TempCompass~\cite{li2025temporal}, FAVOR-Bench~\cite{tu2025favorbench}, and TVBench~\cite{cores2025tvbench}. 

\paragraph{Evaluation Protocol.}
We follow the standard evaluation settings of lmms-eval~\cite{Zhang_2025} and VLMEvalKit~\cite{duan2025vlmevalkitopensourcetoolkitevaluating}, uniformly sampling 16 frames for all short-video benchmarks. All models, including proprietary ones (\textit{e.g.}, GPT-5.2 and Gemini), are evaluated under the same frame budget to ensure a fair comparison.

\subsection{Main Results}

\noindent\textbf{Results on MotionAtlas-Bench.}
Table~\ref{tab:bench_results} reports the performance on our MotionAtlas-Bench under both Single-Frame and Full-Sequence Grounding settings. 
Overall, region-level motion captioning proves highly challenging: even the most capable proprietary models struggle to achieve high recall on fine-grained details. 
In contrast, fine-tuning on MotionAtlas-Data yields consistent and substantial improvements across all base models. 
Moreover, the Full-Sequence setting consistently outperforms Single-Frame Grounding, indicating that continuous spatial prompting effectively strengthens the model's capacity to caption fine-grained motion.


\begin{table}[t]
\centering\small
\caption{Main results on motion-related video understanding benchmarks.
Based on Qwen3-VL-8B~\cite{bai2025qwen3vltechnicalreport}, our MotionAtlas achieves competitive performance even compared with closed-source models, especially on MotionBench~\cite{hong2025motionbench} and TempCompass~\cite{li2025temporal}.
}
\label{tab:main_results}
\vspace{-10pt}
\setlength{\tabcolsep}{1.5pt}
\resizebox{\textwidth}{!}{
\begin{tabular}{l lllllll}
\toprule
\multirow{2}{*}{\textbf{Model}} & \textbf{Motion} &
\textbf{DREAM-1K} &
\multirow{2}{*}{\textbf{TOMATO~}\cite{shangguan2025tomato}} &
\textbf{NExT-QA} &
\textbf{Temp} &
\multirow{2}{*}{\textbf{FAVOR\cite{tu2025favorbench}}} &
\multirow{2}{*}{\textbf{TVBench\cite{cores2025tvbench}}}  \\
& \textbf{Bench}~\cite{hong2025motionbench} & F1-Score~\cite{wang2024tarsierrecipestrainingevaluating} & & MCQ~\cite{xiao2021next} & \textbf{Compass}~\cite{li2025temporal} \\
\midrule
\multicolumn{8}{l}{\textit{Closed-source Models}} \\
GPT-5.2~\cite{openai_gpt5_2} & 65.4 & 42.2 & 53.0 & 79.9 &73.0 & 56.8 & 53.8 \\
Gemini 2.5 Pro~\cite{comanici2025gemini25pushingfrontier} & 62.0 & 42.7 & 48.6 & 79.8 & 73.7	& 58.8 & 59.9 \\
Gemini 3 Pro & 62.6 & 42.5 & 48.3 & 79.6	& 74.1	& 58.5 & 56.8 \\
\midrule
\multicolumn{8}{l}{\textit{Open-source Models}} \\
Molmo2-4B~\cite{clark2026molmo2} &59.9 & 25.7 & 30.3 & 82.4	& 70.2 & 55.8 & 50.6\\
\ + MotionAtlas-Data & 60.6 \up{0.7}& 32.7 \up{7.0} & 35.9 \up{5.6} & 83.6 \up{1.2}	& 75.1 \up{4.9} & 57.4 \up{1.6} &	52.1 \up{1.5}\\
Molmo2-8B~\cite{clark2026molmo2} & 60.5 & 29.6	& 33.8	& 83.6	& 72.1	& 57.9	& 54.1\\
\ + MotionAtlas-Data &61.8 \up{1.3} & 34.0 \up{4.4}& 36.8 \up{3.0}	& 84.3 \up{0.7} & 75.7 \up{3.6}& 58.7 \up{0.8}& 54.8 \up{0.7}\\
\midrule
Qwen3-VL-4B~\cite{bai2025qwen3vltechnicalreport} & 55.9 & 35.6 & 27.4 & 71.6 & 69.6 & 47.0 & 47.2 \\
\ + MotionAtlas-Data & 61.9 \up{6.0} & 38.9 \up{3.3} & 35.2 \up{7.8} & 74.0 \up{2.4} & 74.2 \up{4.6} & 55.0  \up{8.1} & 51.2 \up{4.0}\\
Qwen3-VL-8B~\cite{bai2025qwen3vltechnicalreport} & 59.0	& 38.7 & 34.0 & 76.9 & 71.8 & 54.1 & 51.4 \\
\ + MotionAtlas-Data & 62.6 \up{3.6} & 39.6 \up{0.9} & 36.5 \up{2.5} & 77.2 \up{0.2} & 75.1 \up{3.3} 	& 57.7	\up{3.6} & 52.9 \up{1.5} \\
\bottomrule
\end{tabular}}
\end{table}

\paragraph{Results on General Video Benchmarks.}
Table~\ref{tab:main_results} evaluates all models on 7 public motion-related video understanding benchmarks.
On most benchmarks, proprietary models lead by a moderate margin in temporal and fine-grained motion perception (\textit{e.g.}, TempCompass~\cite{li2025temporal}, FAVOR-Bench~\cite{tu2025favorbench}). 
However, open-source models of comparable size (Molmo2-8B~\cite{clark2026molmo2}, Qwen3-VL-8B~\cite{bai2025qwen3vltechnicalreport}) already achieve competitive or even superior scores on several tasks such as NExT-QA~\cite{xiao2021next}, indicating that the gap largely lies in fine-grained motion perception rather than general video QA.

A key finding is that training on MotionAtlas-Data, which consists entirely of \textit{region-level} motion captions, \textit{consistently} improves performance on \textit{general} (non-region) motion benchmarks. 
For example, Qwen3-VL-4B~\cite{bai2025qwen3vltechnicalreport} gains +6.0 on MotionBench~\cite{hong2025motionbench}, +7.8 on TOMATO~\cite{shangguan2025tomato}, +8.1 on FAVOR-Bench~\cite{tu2025favorbench}, and +4.6 on TempCompass~\cite{li2025temporal}, indicating that learning to describe \textit{fine-grained motion generalizes to broader video tasks}. 
The largest gains arise on benchmarks emphasizing temporal dynamics and action-attribute recognition (TOMATO~\cite{shangguan2025tomato}, FAVOR-Bench~\cite{tu2025favorbench}), consistent with the fine-grained motion coverage of our data. 
Qwen3-VL-8B~\cite{bai2025qwen3vltechnicalreport} and the Molmo-2 series~\cite{clark2026molmo2} follow the same trend, with consistent gains across all benchmarks and no degradation, confirming that our data does \textit{not} induce catastrophic forgetting.

\subsection{Transfer Component Ablation}
\label{sec:transfer_ablation}

To understand \textit{where} this cross-benchmark transfer comes from, we incrementally add each data component and evaluate on general motion benchmarks using Qwen3-VL-4B~\cite{bai2025qwen3vltechnicalreport} (Table~\ref{tab:transfer_ablation}).
Adding general captioning already lifts all three benchmarks, but region-detail captioning yields a markedly larger gain (\textit{e.g.}, TOMATO~\cite{shangguan2025tomato} $27.4\rightarrow31.9$ and FAVOR-Bench~\cite{tu2025favorbench} $47.0\rightarrow55.7$), indicating that the improvements are \textit{not} merely a byproduct of more verbs or generic captioning.
Training with explicit visual region cues further improves transfer to high-dynamic reasoning (TOMATO~\cite{shangguan2025tomato} reaches 35.2, \up{7.8} over the base model).
This suggests that region-grounded supervision teaches models to exploit the temporal density that base models underuse; we provide a supporting analysis showing that \ourdata{} contains cues answerable at 32 frames but not at 16 frames in Appendix~\ref{sec:appendix_fps}.

\subsection{Data Scale Ablation}


\begin{table}[!t]
\centering\small
\caption{
Performance comparisons with varying training data scales.
Model performance continues to improve as the scale of training data increases.
However, these performance improvements are significantly diminished when training proceeds without our proposed MotionAtlas-Data.
}
\label{tab:data_scale}
\vspace{-10pt}
\setlength{\tabcolsep}{3pt}
\resizebox{\textwidth}{!}{
\begin{tabular}{l llllllll}
\toprule
\textbf{Method} & \textbf{Motion} & \textbf{Motion} & \textbf{DREAM-1K} & \multirow{2}{*}{\textbf{TOMATO}~\cite{shangguan2025tomato}} & \textbf{NExT-QA} & \textbf{Temp} & \multirow{2}{*}{\textbf{FAVOR}~\cite{tu2025favorbench}} & \multirow{2}{*}{\textbf{TVBench}~\cite{cores2025tvbench}} \\
(Data Scale) & \textbf{Atlas} & \textbf{Bench}~\cite{hong2025motionbench} & F1-Score~\cite{wang2024tarsierrecipestrainingevaluating} & & MCQ~\cite{xiao2021next} & \textbf{Compass~\cite{li2025temporal}} & & \\
\midrule
Qwen3-VL-4B~\cite{bai2025qwen3vltechnicalreport} & 19.2 & 55.9 & 35.9 & 27.4 & 71.6 & 69.6 & 47.0 & 47.2 \\
\midrule
\multicolumn{9}{l}{\textit{w/} MotionAtlas-Data} \\
~~20\% (32K) & 22.9 & 58.9 & 36.9 & 28.4 & 72.2 & 71.2 & 48.1 & 47.0 \\
~~60\% (95K) & 24.6 & 59.5 & 37.0 & 30.1 & 73.0 & 72.3 & 50.9 & 49.0 \\
~~100\% (159K) & \textbf{28.3} & \textbf{61.9} & \textbf{38.9} & \textbf{35.2} & \textbf{74.0} & \textbf{74.2} & \textbf{55.0} & \textbf{51.2} \\
\midrule
\multicolumn{9}{l}{\textit{w/o} MotionAtlas-Data} \\
~~20\% (32K) & 12.9 & 57.4 & 37.3 & 29.0 & 70.9 & 70.8 & 47.4 & 46.7 \\
~~60\% (95K) & 12.9 & 58.8 & 36.9 & 30.7 & 71.3 & 72.3 & 50.6 & 47.4 \\
~~100\% (159K) & 12.2 & 60.5 & 38.3 & 28.4 & 71.9 & 73.3 & 52.2 & 48.5 \\
\bottomrule
\end{tabular}
}
\end{table}


\begin{figure*}[!t] 
    \centering
    \begin{minipage}[c]{0.4\textwidth}
        \centering
        \includegraphics[width=\linewidth]{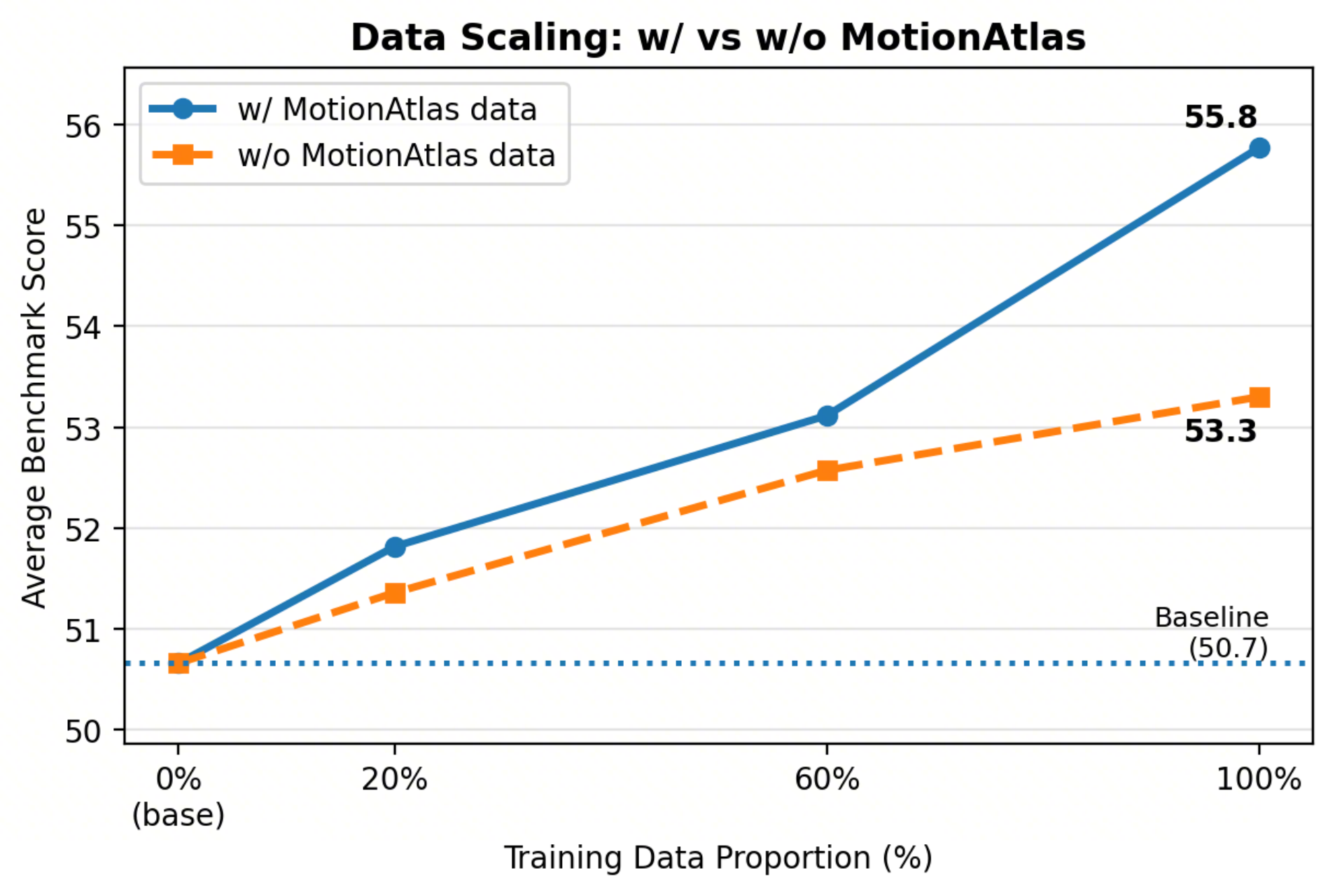}
        \vspace{-20pt}
        \caption{Data scaling curves.
        Adding MotionAtlas-Data brings more significant improvements.
        }
        \label{fig:data_scale}
    \end{minipage}\hfill
    \begin{minipage}[c]{0.58\textwidth}
        \centering\small
        \makeatletter\def\@captype{table}\makeatother 
        \caption{Ablation of the MotionAtlas data pipeline (MA Pipeline) components, evaluated on MotionAtlas-Bench via caption quality judging.}
        \label{tab:pipeline_ablation}
        \vspace{-10pt}
        \setlength{\tabcolsep}{4pt} 
        \renewcommand{\arraystretch}{1.5} 
        \resizebox{\linewidth}{!}{
        \begin{tabular}{l ccc}
        \toprule
        Method & Acc ($\uparrow$) & Recall ($\uparrow$) & Precision ($\uparrow$)\\
        \midrule
        MA Pipeline (full) & \textbf{39.9} & \textbf{68.2} & \textbf{58.5} \\
        \midrule
        ~~w/o Self-Bootstrap & 36.4 & 64.1 & 56.8 \\
        ~~w/o Full-Video Caption & 33.2 & 58.9 & 56.4 \\
        ~~w/o Spatial Crop & 32.7 & 60.9 & 53.6 \\
        \bottomrule
        \end{tabular}}
    \end{minipage}
\end{figure*}

Using Qwen3-VL-4B~\cite{bai2025qwen3vltechnicalreport} as the base model, Table~\ref{tab:data_scale} compares training with and without MotionAtlas-Data at three data scales (32K, 95K, 159K).

\paragraph{Scaling with MotionAtlas-Data.}
When MotionAtlas-Data is included, performance scales monotonically across all benchmarks: MotionAtlas-Bench rises from 22.9 (32K) to 28.3 (159K), a +5.4 gain, and external benchmarks follow the same trend (TOMATO~\cite{shangguan2025tomato} $28.4\rightarrow35.2$, FAVOR-Bench~\cite{tu2025favorbench} $48.1\rightarrow55.0$).

\paragraph{Necessity of MotionAtlas-Data.}
In contrast, replacing MotionAtlas-Data with an equal amount of TarsierRecap data~\cite{yuan2025tarsier2} yields essentially no improvement on MotionAtlas-Bench ($\sim$12--13\% at all scales), showing that its \textit{distinctive fine-grained details} have no viable substitute. 
External benchmarks may still improve without our data, but the gains are consistently \textit{smaller} (\textit{e.g.}, $37.3\rightarrow38.3$ \textit{vs.}\ $36.9\rightarrow38.9$ on DREAM-1K~\cite{wang2024tarsierrecipestrainingevaluating}), further validating its complementary value.

\subsection{Pipeline Component Ablation}

To isolate the contribution of each component in our MotionAtlas (MA) data pipeline (\S\ref{sec:scbs}), we ablate three key designs, self-bootstrap refinement, full-video captioning, and spatial cropping, and evaluate the resulting annotation quality on MotionAtlas-Bench (Table~\ref{tab:pipeline_ablation}).

\paragraph{Self-Bootstrap Refinement.}
Removing the dual-rollout self-bootstrap stage leads to a $-$3.5 accuracy drop (39.9 $\to$ 36.4). 
Without this module, single-pass VLM descriptions retain more uncorrected hallucinations, raising error rates and reducing the model's tendency to describe fine-grained details.

\paragraph{Full-Video Caption.}
Relying on local captions only causes the largest accuracy decline ($-$6.7, to 33.2). 
The global caption resolves temporal inconsistencies between adjacent events, removes speculative descriptions at event boundaries, and provides a coherent timeline that anchors the local narratives.

\paragraph{Spatial Crop.}
Feeding only the original video segments, without spatial zoom-in, drops accuracy by $-$7.2 (to 32.7). 
This confirms that cropping to the target entity's bounding box is essential for directing the VLM's attention to the relevant motion and away from background clutter and other moving objects.

\subsection{Qualitative Results}

\begin{figure}[t]
\centering
\includegraphics[width=\linewidth]{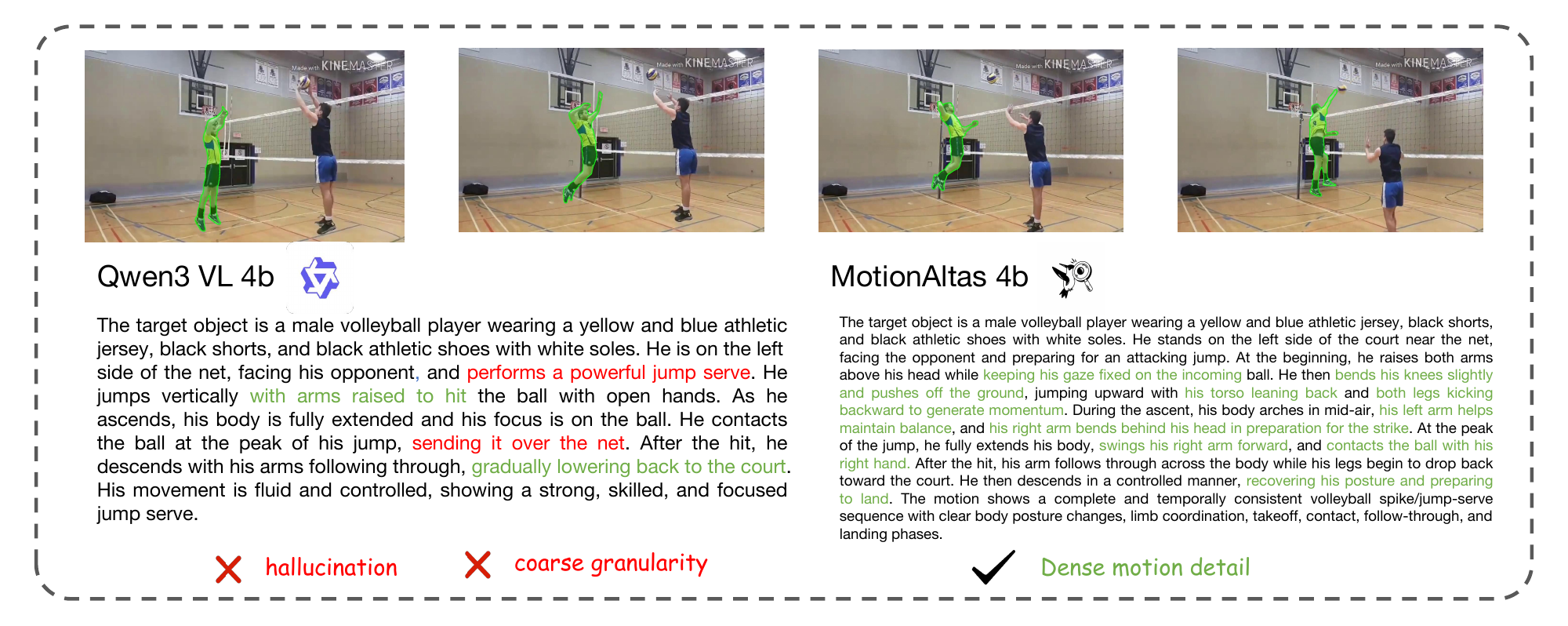}
\caption{Qualitative comparison between different datasets.}
\label{fig:data_compare}
\end{figure}

Figure~\ref{fig:data_compare} compares Qwen3-VL-4B with MotionAtlas-4B.
The base model tends to produce coarse motion descriptions and occasionally hallucinated events, whereas MotionAtlas-4B generates denser and more precise descriptions that capture fine-grained body dynamics throughout the action.
\section{Related Work}
\label{sec:related_work}
\paragraph{MLLMs for video understanding.}
Driven by large-scale datasets and architectural advancements, MLLMs have achieved remarkable progress in general video understanding~\cite{wang2025internvl3, clark2026molmo2, bai2025qwen3vltechnicalreport}.
Recent foundation models (e.g., Qwen3-VL~\cite{bai2025qwen3vltechnicalreport}, InternVL3.5~\cite{wang2025internvl3}, Molmo2~\cite{clark2026molmo2}, VideoLLaMA 3~\cite{zhang2025videollama3frontiermultimodal}, InternVideo2.5~\cite{wang2025internvideo25empoweringvideomllms}) and instruction-tuning efforts (e.g., LLaVA-Video~\cite{zhang2024llava}, LLaVA-OneVision-1.5~\cite{an2025llava}) demonstrate strong capabilities in temporal scene comprehension and user alignment.
However, while excelling globally, these models still struggle with fine-grained motion dynamics and precise spatiotemporal reasoning over localized regions~\cite{zhang2025videollama3frontiermultimodal,wang2025internvideo25empoweringvideomllms,meng2025openo3}.

\paragraph{Video motion understanding.}
High-quality datasets are a primary driver of multimodal foundation models~\cite{zhang2024llava, chen2024sharegptvideo, li2025denseworld1m}, yet large-scale resources for motion understanding remain scarce. 
Traditional benchmarks predominantly target global action recognition or coarse temporal reasoning (\eg, ActivityNet-QA~\cite{yu2019activityqa}, NExT-QA~\cite{xiao2021next}, MVBench~\cite{li2024mvbench}, Video-MME~\cite{fu2025video}). 
Unlike atomic or fine-grained action recognition, which primarily asks what action occurs at the clip level, MotionAtlas evaluates how a specified region moves, covering trajectory, part articulation, interaction, state change, and camera-related motion under explicit spatial grounding.
To push towards detailed motion perception, recent works like MotionBench~\cite{hong2025motionbench}, Tarsier~\cite{wang2024tarsierrecipestrainingevaluating,yuan2025tarsier2}, MotionSight~\cite{du2025motionsight}, FAVOR-Bench~\cite{tu2025favorbench}, and TVBench~\cite{cores2025tvbench} introduce fine-grained motion annotations. 
Despite this, global motion captioning still suffers from evaluation intractability and severe hallucinations~\cite{chai2025auroracap,chen2024sharegptvideo}, complicating data quality maintenance and faithfulness verification.

\paragraph{Region-level detailed understanding.}
Concurrently, region-level captioning has proven effective for describing detailed objects and their attributes~\cite{chen2023shikraunleashingmultimodalllms, yuan2024osprey, guo2024regiongpt, peng2024vasttrack, wang2025grasp, OMGLLaVA}, with recent works extending this to spatial-temporal video domains~\cite{yuan2025videorefer, yuan2025pixelreferunifiedframeworkspatiotemporal, wang2025spatialvid, athar2025vicas, sa2va}. 
However, these methods predominantly address visual appearance and static attributes, leaving region-level fine-grained motion dynamics largely unexplored. 
MotionAtlas bridges this gap by confining motion descriptions to region-level spatiotemporal masks, decoupling visual clutter from motion entanglement, and thereby establishing a scalable, hallucination-resistant paradigm for both data generation and reliable evaluation.

\section{Conclusion}
\label{sec:conclusion}


In this work, we presented \textbf{MotionAtlas}, a framework that systematically addresses region-level motion understanding in Video-MLLMs.
Recognizing the evaluation intractability of global video captioning, we shift the focus to region-level reasoning and introduce \textbf{MotionAtlas-Bench}, a diagnostic benchmark that evaluates event-centric, region-grounded motion through a reliable checklist-style MCQ protocol with explicit identity references, temporal anchors, and multi-aspect attributes.
We further proposed a rigorous and scalable data pipeline that combines temporal segmentation, VLM-based self-difference judgment, and multi-source summarization, yielding \textbf{MotionAtlas-Data}, a large-scale corpus with exceptionally dense action verbs and minimal hallucinations.
Powered by this data and a tailored training recipe, our Video-MLLMs achieve consistent and significant improvements over strong baselines on both motion-specific and general video understanding tasks. 
We believe MotionAtlas establishes a solid foundation for fine-grained video motion understanding and points toward more robust and precise video-language modeling.

\paragraph{Limitations and future work.}
MotionAtlas currently targets the single-target setting; extending our region-grounded formulation to multi-object scenarios with interactions and identity correspondence is a natural next step.
Moreover, the large VLM backbone is used only for offline data construction, while deployment relies on the distilled MotionAtlas-4B/8B models.

\bibliographystyle{splncs04}
\bibliography{main}

\clearpage
\appendix
\noindent
\textbf{Overview.} In this appendix, we provide additional implementation details, qualitative results, and analyses to complement the main paper. Furthermore, we include a \textbf{supplementary demo video} that dynamically illustrates our data curation pipeline and showcases qualitative results of fine-grained motion understanding. The remainder of this appendix is organized as follows:

\begin{itemize}
    \item \textbf{Section~\ref{sec:appendix_benchmark}} presents full details of our benchmark construction, including data curation (Sec.~\ref{sec:appendix_curation}), human refinement protocols (Sec.~\ref{sec:appendix_event} and \ref{sec:appendix_mcq_refine}), annotation reliability (Sec.~\ref{sec:appendix_reliability}), benchmark robustness and diversity (Sec.~\ref{sec:appendix_robustness}), as well as benchmark statistics and qualitative cases.
    \item \textbf{Section~\ref{sec:appendix_pipeline}} provides complete technical details of the dataset construction pipeline, covering the self-bootstrap refinement formulas (Sec.~\ref{sec:appendix_bootstrap}), multi-source narrative synthesis strategy (Sec.~\ref{sec:appendix_synthesis}), and entity-centric motion score derivation (Sec.~\ref{sec:appendix_motion_score}).
    \item \textbf{Section~\ref{sec:appendix_exp}} reports supplementary experiments, specifically focusing on data source effectiveness (Sec.~\ref{sec:appendix_data_source}), judge robustness (Sec.~\ref{sec:appendix_judge}), and frame sampling strategies (Sec.~\ref{sec:appendix_fps}).
    \item \textbf{Section~\ref{sec:ethics}} discusses the broader impacts and ethical considerations of our work.
\end{itemize}

\section{Benchmark Details}
\label{sec:appendix_benchmark}

\subsection{Data Curation Details}
\label{sec:appendix_curation}
To construct a cross-domain benchmark for fine-grained referential motion, we assemble a 600-sample pool equally drawn from DREAM-1K~\cite{wang2024tarsierrecipestrainingevaluating} (video events), MeViS~\cite{ding2023mevis} (native masks), and MotionBenchCaption~\cite{hong2025motionbench} (motion captions). We sample 16 frames per candidate, discarding clips under 480p or shorter than 3 seconds. Using Gemini 3 Pro, we assess motion attributes (including multiple atomic motions, multi-object interactions, and small/background objects) to select 166 medium/hard cases. We supplement this with 26 FineDiving~\cite{xu2022finediving} (action labels/segments) and 6 MotionBenchQA~\cite{hong2025motionbench} (temporal QA pairs) samples, yielding 198 candidates.
Following human review, we finalized 107 benchmark videos. For non-MeViS sources, instance masks are unified via an interactive SAM2~\cite{ravi2024sam}-assisted workflow.

\subsection{Human Refinement for Event Annotation}
\label{sec:appendix_event}
\begin{figure}[t]
    \centering
    \includegraphics[width=\textwidth]{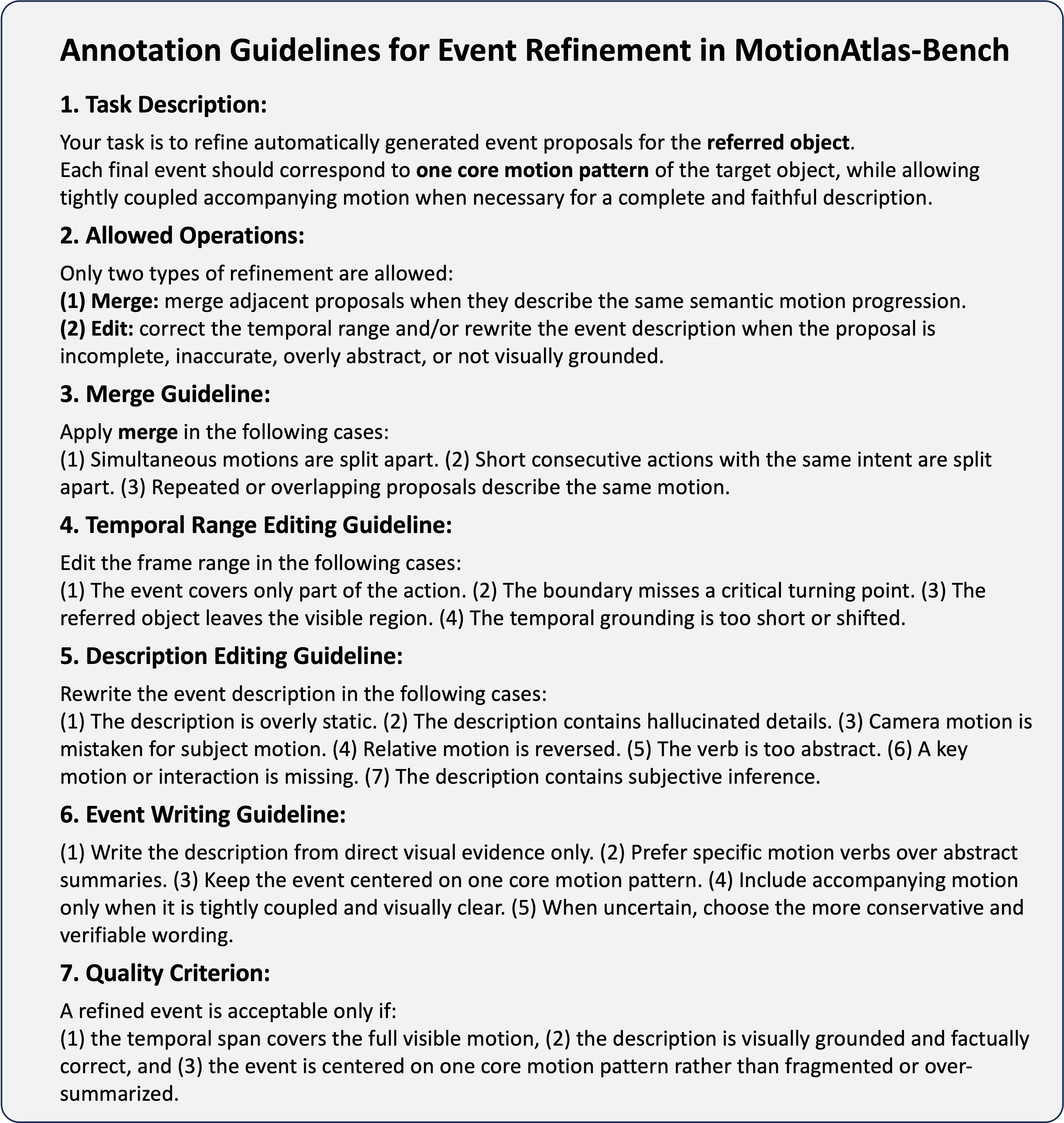}
    \caption{Human refinement guidelines for event segmentation and description. It illustrates the core principles for merging adjacent proposals and correcting boundary shifts.}
    \label{fig:anno_guide_event}
\end{figure}

While Gemini 3 Pro proposals provide initialization, they often over-segment motion, truncate temporal spans, or hallucinate details. To ensure each event captures exactly one core motion pattern (inclusive of accompanying motions), annotators perform manual refinement. As illustrated in Figure~\ref{fig:anno_guide_event}, adjacent proposals with identical semantic intent are merged. Furthermore, incomplete boundaries, reversed relative motions, or ungrounded speculations are manually corrected. Final descriptions strictly rely on observable evidence.

\subsection{MCQ Option Construction}
\label{sec:appendix_mcq_construction}
We decompose dense event descriptions into atomic facts, where each fact corresponds to one independently verifiable motion attribute: $f=(s,a,t,m,r,v)$, where $s$ is the subject, $a$ is the core action, $t$ is the temporal anchor, $m$ is the motion aspect, $r$ is the queried attribute, and $v$ is the correct value. To rigorously evaluate fine-grained understanding and penalize model hallucinations, each MCQ (6--10 options) generated by Gemini 3 Pro intentionally combines fact-preserving answers, challenging distractors, and diagnostic negatives. Specifically, the options include: (i) one exact correct value; (ii) several hard distractors drawn from the same attribute space, differing from the correct answer by only one visually confusable detail to test precise visual discrimination; and (iii) three special negative options designed for distinct diagnostic purposes: \textit{not mentioned} and \textit{mentioned but no value} serve to penalize missed observations and explicitly assess recall, whereas \textit{mentioned but different value} acts as a fallback to detect ungrounded hallucinations. Finally, we ensure all special negatives are present, remove duplicates, and deterministically shuffle the final order.

\subsection{Human Refinement for MCQ Verification}
\label{sec:appendix_mcq_refine}
\begin{figure}[t]
    \centering
    \includegraphics[width=\textwidth]{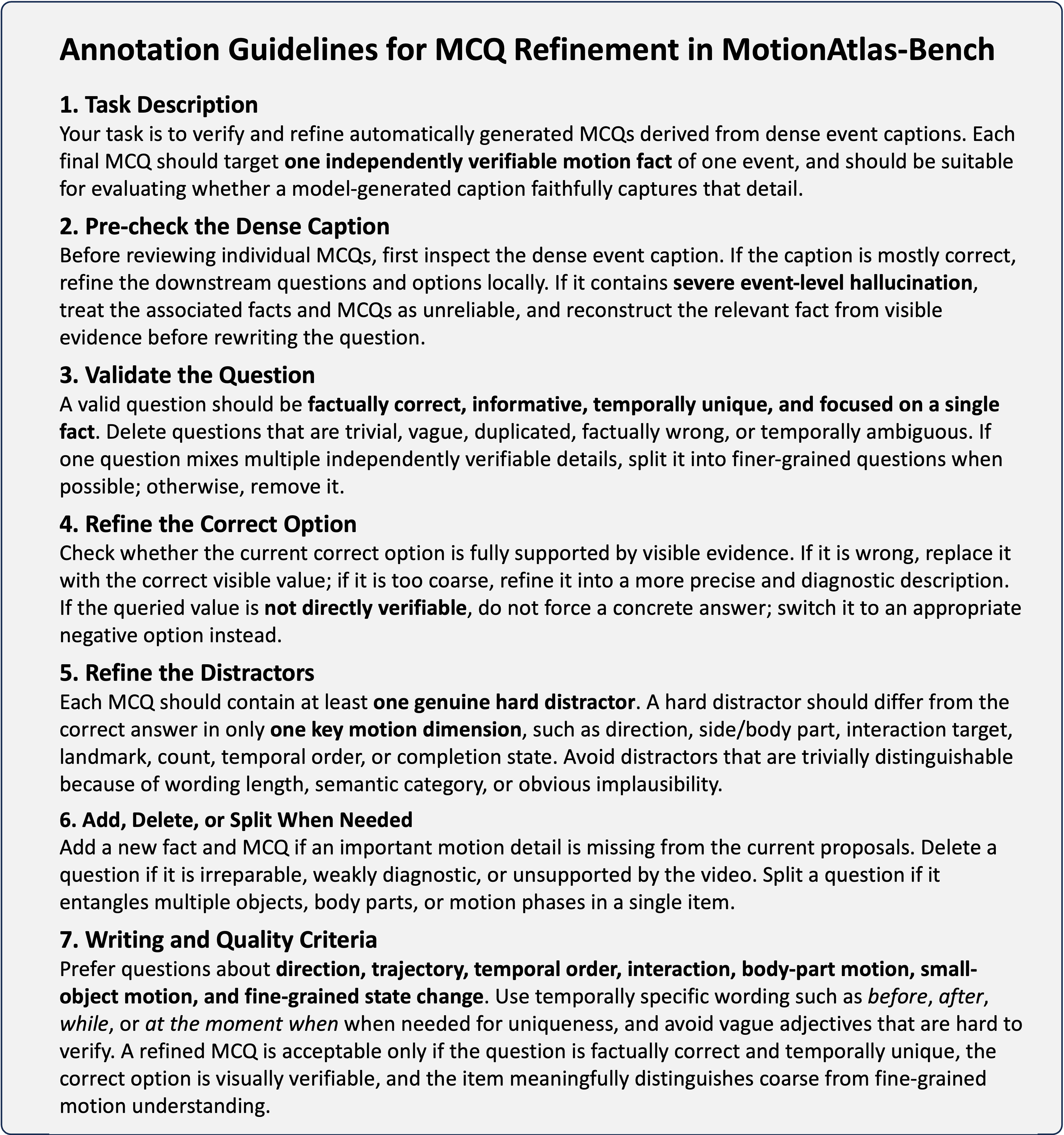}
    \caption{Quality control guidelines for MCQ factual verification. The figure illustrates the standard rules for assessing option validity and hard distractor discriminability.}
    \label{fig:anno_guide_mcq}
\end{figure}

Since VLM hallucinations can propagate into option design, we conduct manual verification, leveraging Gemini 3 Pro scoring strictly as a prioritization cue. As shown in Figure~\ref{fig:anno_guide_mcq}, human refinement ensures the correct option matches directly observable video motion and validates that at least one distractor differs by only one key attribute to form a genuine hard case. Missing details trigger manual fact and MCQ creation, while unverified queried values are replaced with negative options. Unreliable questions are removed, and entangled multi-detail questions are split into independent attributes.

\subsection{Annotation Reliability}
\label{sec:appendix_reliability}

To quantify the reliability of our human-in-the-loop annotation, we report both inter-annotator agreement and the statistics of the human-refinement process.

\begin{table}[t]
\centering\small
\caption{Inter-annotator agreement (IAA) from a pilot study. We report raw agreement (Agr.) and Cohen's $\kappa$. The final MCQ answer is highly reliable, while finer-grained aspect labeling is inherently more subjective.}
\label{tab:iaa}
\vspace{-10pt}
\setlength{\tabcolsep}{8pt}
\begin{tabular}{lrr}
\toprule
Aspect & Agr. & $\kappa$ \\
\midrule
Final MCQ answer & 0.91 & 0.90 \\
Fact validity    & 0.89 & 0.62 \\
Aspect label     & 0.63 & 0.47 \\
\bottomrule
\end{tabular}
\end{table}

\begin{table}[t]
\centering\small
\caption{Human-refinement funnel for events and MCQs. Starting from raw VLM proposals, a large fraction is deleted, and most of the kept items are further revised by human annotators.}
\label{tab:refine_funnel}
\vspace{-10pt}
\setlength{\tabcolsep}{5pt}
\resizebox{\textwidth}{!}{
\begin{tabular}{lrrrr}
\toprule
Item & Raw & Deleted & Final kept & Human revised \\
\midrule
Events & 449 (100\%) & 177 (39.4\%) & 272 (60.6\%) & 370 (82.4\%) \\
MCQs   & 3{,}120 (100\%) & 1{,}047 (33.6\%) & 2{,}073 (66.4\%) & 1{,}457 (70.3\%) \\
\bottomrule
\end{tabular}}
\end{table}

\noindent\textbf{Inter-Annotator Agreement.} As shown in Table~\ref{tab:iaa}, the final MCQ answer reaches $0.91$ raw agreement ($\kappa=0.90$), and fact validity reaches $0.89$ agreement, confirming that the core benchmark labels are highly reliable. The lower agreement on fine-grained aspect labels ($0.63$) reflects the inherent subjectivity of categorizing motion aspects, which is why we treat aspect-level slices as diagnostic rather than standalone claims.

\noindent\textbf{Human-Refinement Funnel.} As reported in Table~\ref{tab:refine_funnel}, every MCQ is double-checked by humans. Of the raw VLM-proposed MCQs, $33.6\%$ are deleted, and $70.3\%$ of the final kept MCQs are field-refined; only $29$ of the $2{,}073$ final MCQs are added from scratch by humans. This indicates that the VLM proposals primarily serve to direct human attention to dense motion details, while the final labels are firmly grounded in human verification.

\subsection{Benchmark Statistics}
\label{sec:appendix_benchmark_statistics}

\begin{figure}[t]
    \centering
    \begin{subfigure}[b]{0.32\textwidth}
        \centering
        \includegraphics[width=\textwidth]{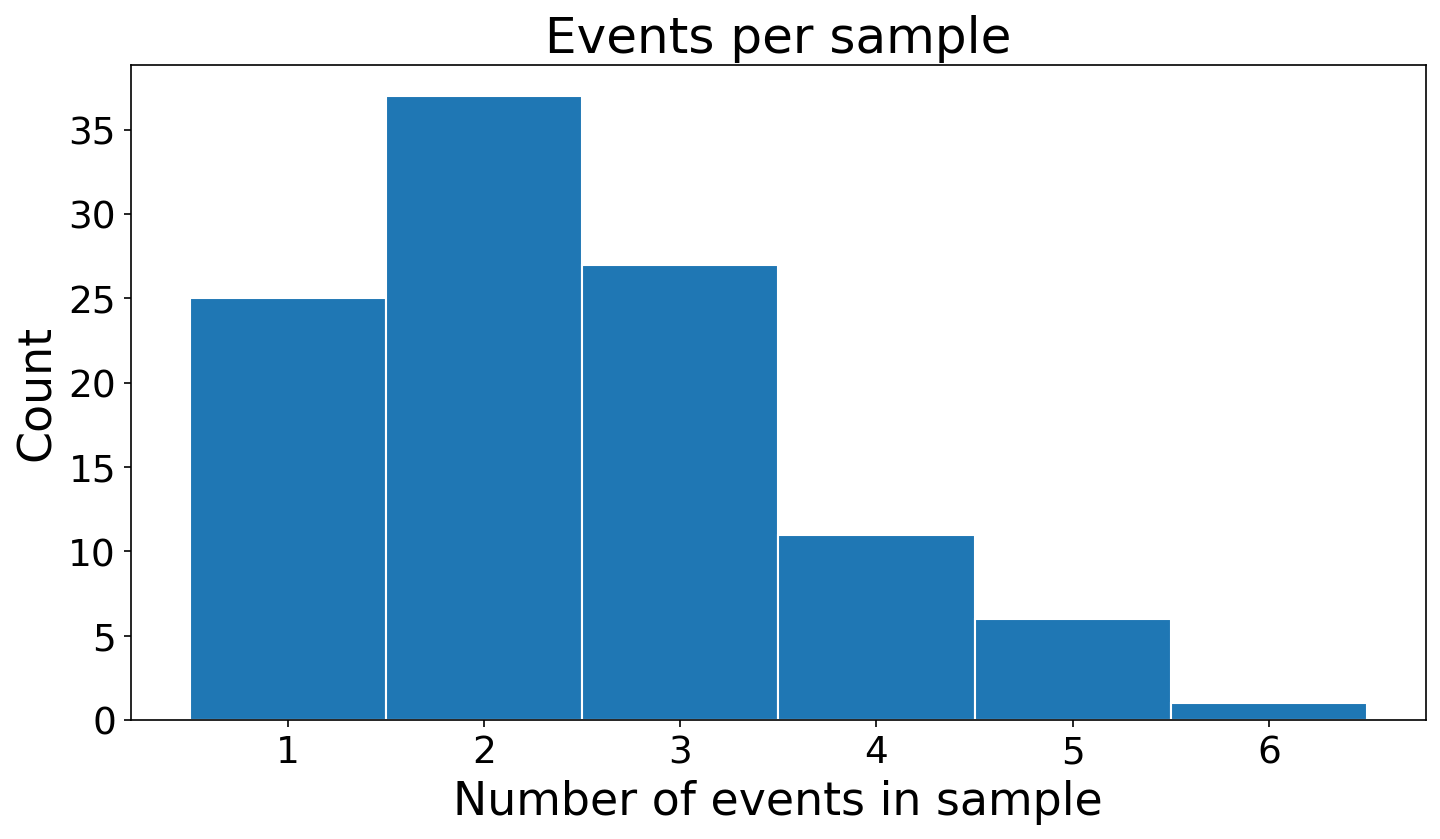}
        \caption{Events per Sample}
        \label{fig:events_per_sample}
    \end{subfigure}
    \hfill
    \begin{subfigure}[b]{0.32\textwidth}
        \centering
        \includegraphics[width=\textwidth]{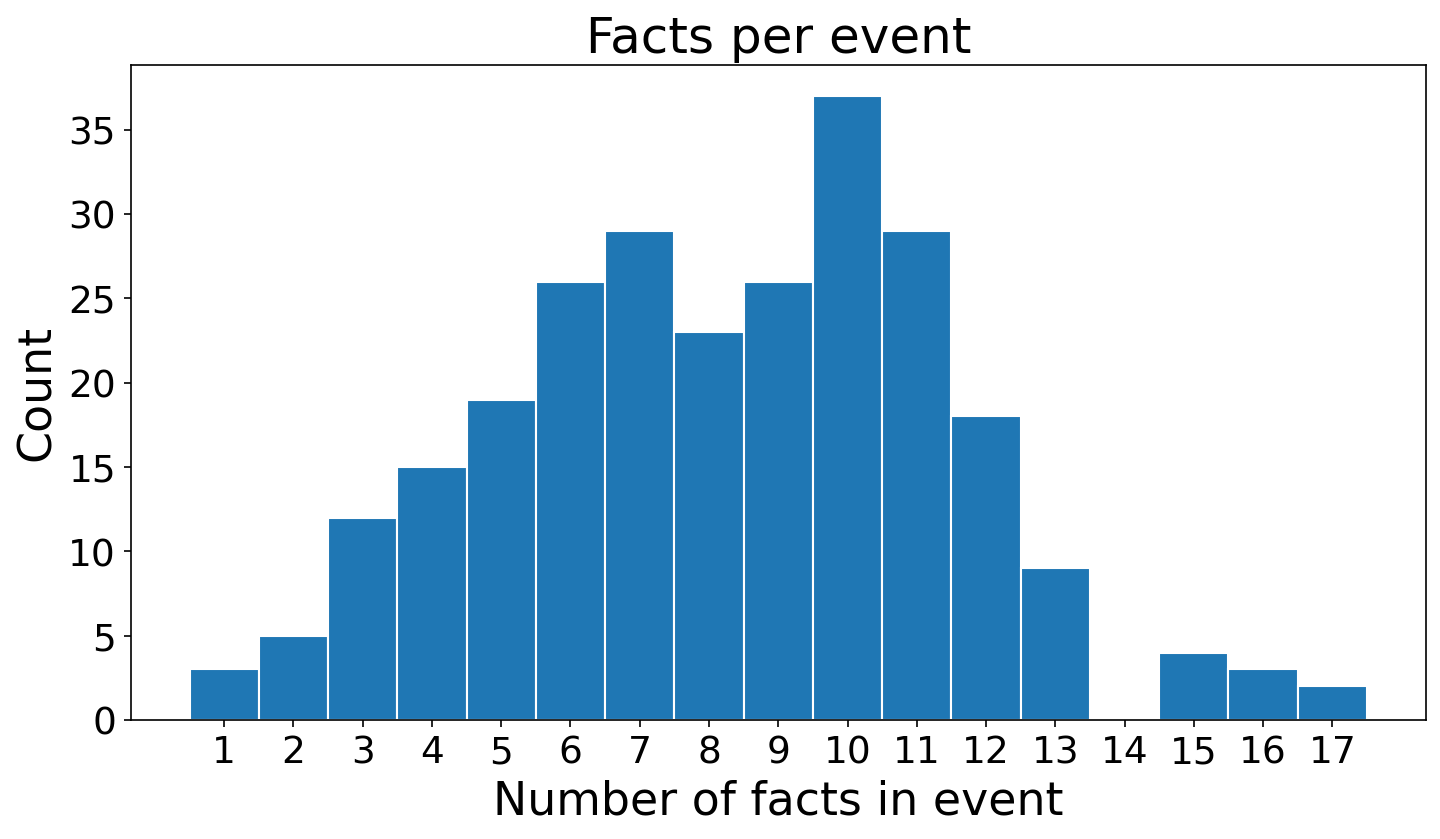}
        \caption{Facts per Event}
        \label{fig:facts_per_event}
    \end{subfigure}
    \hfill
    \begin{subfigure}[b]{0.32\textwidth}
        \centering
        \includegraphics[width=\textwidth]{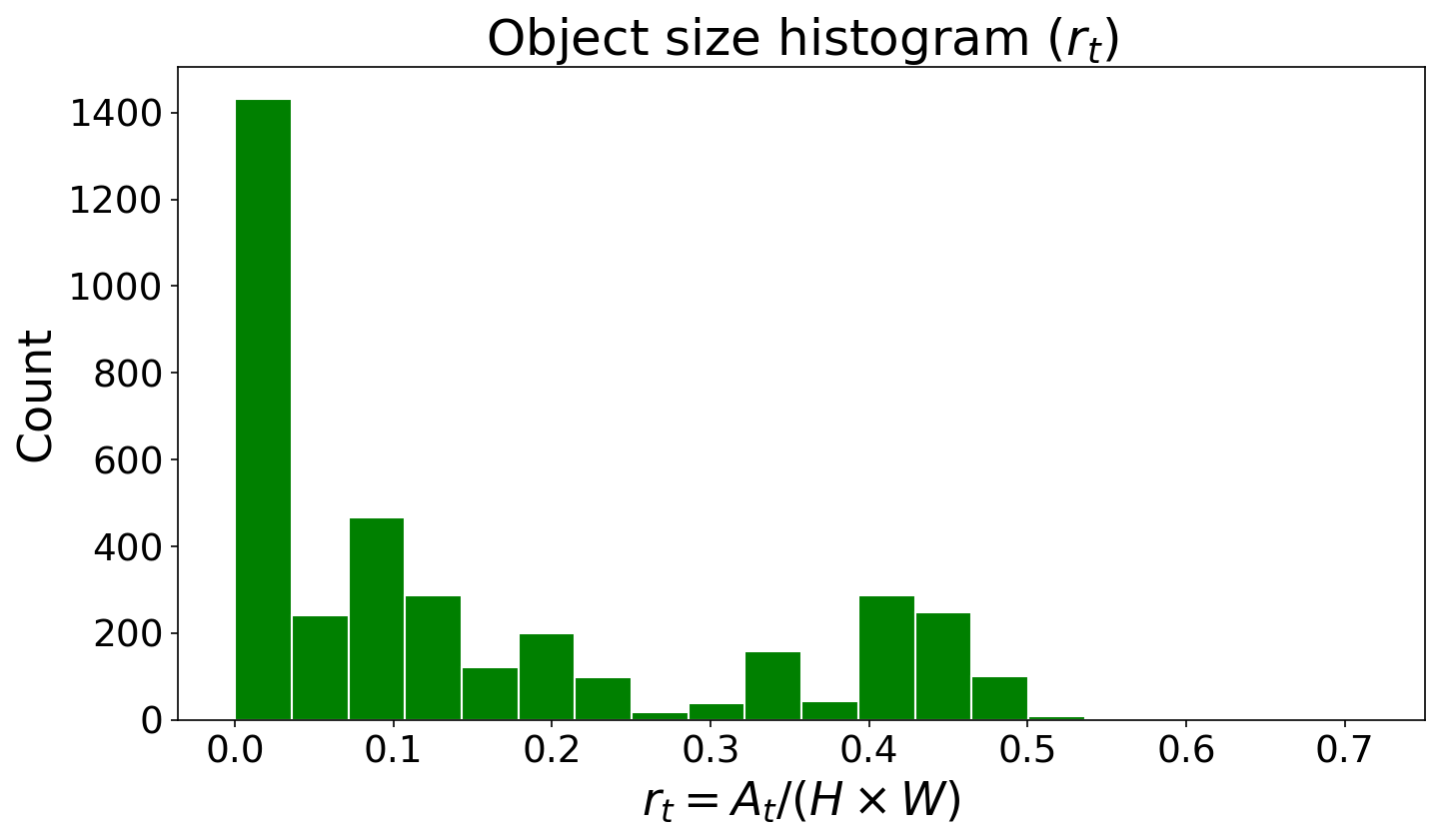}
        \caption{Object Size ($r_t$)}
        \label{fig:rt_distribution}
    \end{subfigure}
    \caption{Data distributions in our benchmark, illustrating the number of events per sample, facts per event, and refer object size ($r_t$).}
    \label{fig:benchmark_statistics}
\end{figure}

\noindent\textbf{Distribution of Number of Events.} As illustrated in Figure~\ref{fig:events_per_sample}, our benchmark features complex multi-event structures. It even contains specific samples with 5 and 6 events, respectively, leading to an advanced requirement of modeling complex temporal relationships and transitions between multiple events.

\noindent\textbf{Distribution of Number of Facts.} As shown in Figure~\ref{fig:facts_per_event}, we provide dense factual annotations within each temporal segment. The benchmark contains on average 8 facts per event, with some extreme cases comprising up to 17 facts. This dense distribution highlights the comprehensive and fine-grained nature of our factual descriptions.

\noindent\textbf{Distribution of Refer Object Size ($r_t$).} We compute the relative area of each reference object mask across a total of 16,559 frame-level mask points. As presented in Figure~\ref{fig:rt_distribution}, the majority of the tracked objects are extremely small, with a sharp peak near 0.0. The mean area across all mask points is roughly 15.1\%. This distribution highlights the importance of addressing small-scale objects and fine-grained spatial understanding in complex videos.

\subsection{Benchmark Robustness and Diversity}
\label{sec:appendix_robustness}

\noindent\textbf{Domain Diversity.} Although the benchmark contains 107 videos, they span diverse domains: sports ($26.2\%$), daily activities ($37.4\%$), animals ($19.6\%$), and object ($16.8\%$). We evaluate $2{,}073$ human-reviewed MCQs in total rather than only the 107 videos, so the effective measurement granularity is far higher than the video count alone suggests.

\noindent\textbf{Full-Set Stability.} Repeated full-benchmark evaluations have small variance (std: Gemini 3 Pro $0.19$, Qwen3-32B $0.17$, Qwen3-4B $0.85$), so the large gaps between models are robust to evaluation noise.

\begin{table}[t]
\centering\small
\caption{Reduced-set ranking robustness. For each reduced subset, we score all models, rank them, and compute Spearman $\rho$ against the full-benchmark ranking, repeating each setting $N{=}5$ times (mean$\pm$std). Rankings remain highly stable ($\rho>0.90$) even at 1/4 scale.}
\label{tab:reduced_set}
\vspace{-10pt}
\setlength{\tabcolsep}{5pt}
\resizebox{\textwidth}{!}{
\begin{tabular}{lcccc}
\toprule
Ratio & Random (MCQ) & Random (video) & Domain-balanced & Aspect-balanced \\
\midrule
$\rho$ @ 3/4 & $0.99{\pm}0.01$ & $0.97{\pm}0.02$ & $0.98{\pm}0.01$ & $0.98{\pm}0.02$ \\
$\rho$ @ 1/2 & $0.97{\pm}0.01$ & $0.93{\pm}0.03$ & $0.95{\pm}0.02$ & $0.95{\pm}0.03$ \\
$\rho$ @ 1/4 & $0.92{\pm}0.02$ & $0.87{\pm}0.05$ & $0.90{\pm}0.04$ & $0.89{\pm}0.06$ \\
\bottomrule
\end{tabular}}
\end{table}

\noindent\textbf{Reduced-Set Ranking Stability.} To verify that the benchmark ranking is not an artifact of its size, we re-score and re-rank all models on reduced subsets and compute Spearman $\rho$ against the full-benchmark ranking (Table~\ref{tab:reduced_set}). Random MCQ subsets remain highly stable even at 1/4 scale ($\rho=0.92$), while video-level subsets stay stable at 1/2 scale ($\rho=0.93$; 1/4 stress test: $\rho=0.87$). Domain-balanced video sampling and aspect-balanced MCQ sampling yield similarly high correlations, confirming that the model ranking is robust to both subset size and distribution bias.

\subsection{Benchmark Cases}
\label{sec:appendix_benchmark_cases}

\begin{figure}[t]
    \centering
    \includegraphics[width=\textwidth]{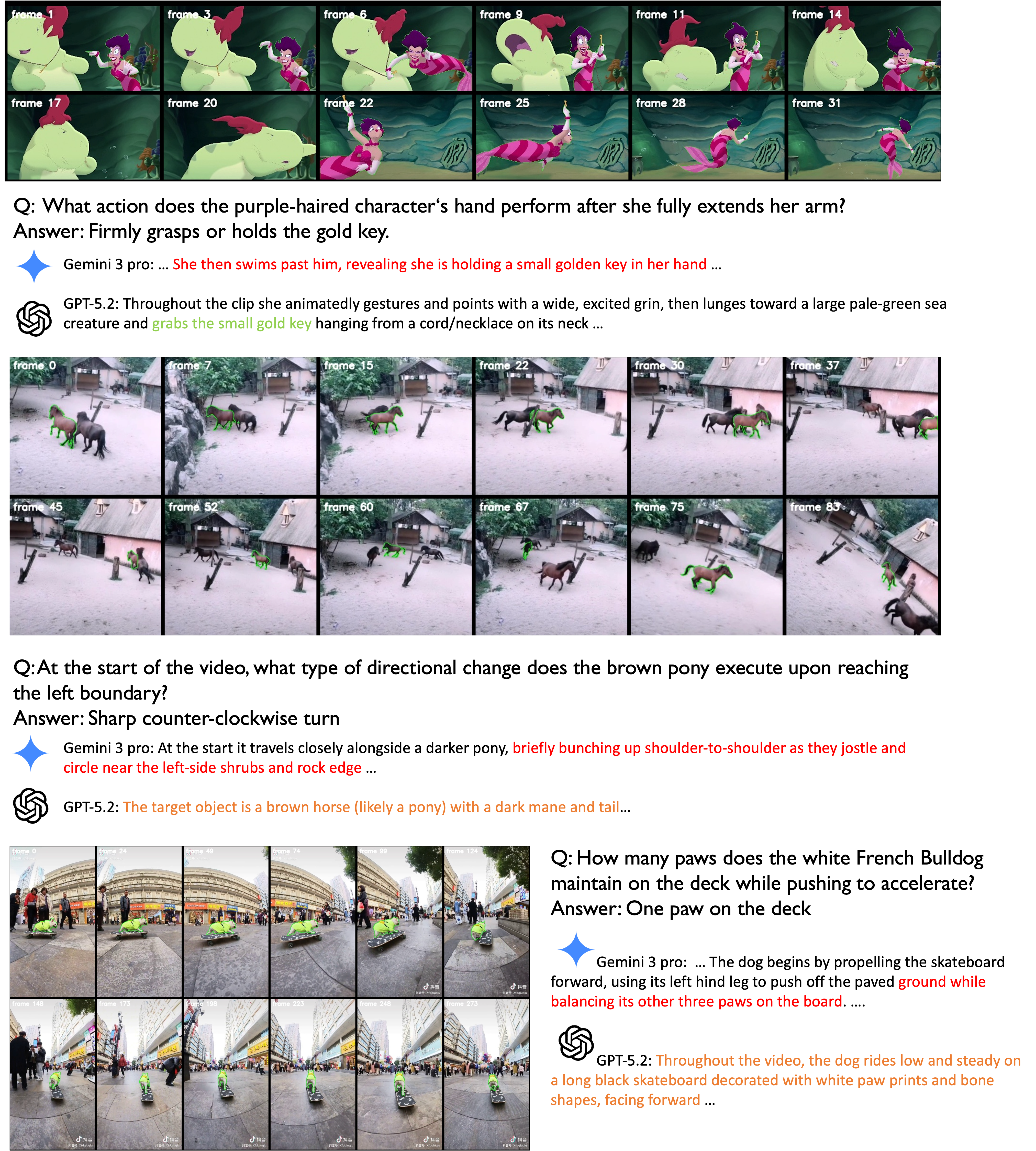}
    \caption{Qualitative benchmark examples illustrating two types of model failures. For each case, we show uniformly sampled frames with the referent object highlighted, the MCQ question and ground-truth answer, and caption predictions from two baseline models. \textbf{Yellow text} denotes a \textit{missed} motion attribute (the detail is entirely absent from the caption); \textbf{red text} denotes a \textit{wrong} description (the attribute is mentioned, but its value is incorrect).}
    \label{fig:appendix_benchmark_cases}
\end{figure}

We present three representative benchmark cases in Figure~\ref{fig:appendix_benchmark_cases} to illustrate the two principal failure modes of current Video-MLLMs on fine-grained motion attributes. \textbf{Yellow highlights} mark extit{missed} motion details---the model's caption entirely omits a key attribute that is clearly observable in the video. \textbf{Red highlights} mark \textit{wrong} descriptions---the model does mention the relevant motion, but the stated value or property is factually incorrect. Both failure modes are prevalent even in state-of-the-art models, confirming that capturing the precise spatiotemporal details probed by \ourbench{} remains a substantial open challenge.

\section{Dataset Details}
\label{sec:appendix_pipeline}

\subsection{Self-Bootstrap Refinement: Full Formulas}
\label{sec:appendix_bootstrap}

\noindent\textbf{Phase 1: Dual Rollout.}
For each event $E_k$, independently generate a second description, injecting the first-pass descriptions of adjacent events as temporal context:
\begin{equation}
  c_k^{(2)} = M_{\text{cap}}(F_k, \text{FocalCrop}(F_k, \{m_t\}), e, d_k^{\text{seg}}, c_{k-1}^{(1)}, c_{k+1}^{(1)})
\end{equation}

\noindent\textbf{Phase 2: Differential Claim Extraction.}
Extract the set of conflicting claims:
\begin{equation}
  Q_k = \text{Diff}(c_k^{(1)}, c_k^{(2)}) = \{(\text{topic}_j, a_j, b_j, \tilde{c}_j)\}_{j=1}^{J}
\end{equation}

\noindent\textbf{Phase 3: Visual Grounding Judgment.}
3-way blind adjudication:
\begin{equation}
  v_j = M_{\text{judge}}(\sigma(a_j, b_j, \tilde{c}_j), F_k, \text{topic}_j), \quad v_j \in \{A, B, \text{Tie}, \text{None}\}
\end{equation}
where $\sigma$ denotes a random permutation. The distractor serves as a calibration anchor: if the judge selects it, the judgment is deemed unreliable and discarded to prevent error propagation.

\noindent\textbf{Phase 4: Caption Correction.}
\begin{equation}
  \hat{c}_k = \text{Merge}(c_k^{(1)}, \{(v_j, \text{fact}_j)\}_{j=1}^{J})
\end{equation}
This process can be iterated for $R$ rounds.

\subsection{Multi-Source Narrative Synthesis: Strategy}
\label{sec:appendix_synthesis}

The synthesis stage integrates local captions $\{\hat{c}_k\}$ with the full-video caption $c_{\text{full}}$, resolving conflicts by the following priority:
\begin{equation}
  \text{Priority}: \underbrace{\hat{c}_k}_{\text{motion detail}} > \underbrace{c_{\text{full}}}_{\text{scene context}}
\end{equation}
Specific strategies include: (1)~temporal ordering using $c_{\text{full}}$'s global timeline and each event's frame range $[s_k, e_k]$; (2)~removal of speculative descriptions at event boundaries; (3)~deduplication of redundant content shared between adjacent events; (4)~source-priority conflict resolution. The final output is structured into multiple granularities: a brief caption (an overall summary), an event caption (a per-event semantic action phrase), and a detailed narrative (a complete, fine-grained motion narrative).

\subsection{Entity-Centric Motion Score: Full Derivation}
\label{sec:appendix_motion_score}

Given video frame $t$ and the refer entity's mask $m_t$, first compute the optical flow magnitude map:
\begin{equation}
  m_t(\mathbf{x}) = |\mathbf{u}_t(\mathbf{x})|_2
\end{equation}
Define three spatial regions: $\Omega_{\text{core}}$ (mask interior), $\Omega_{\text{ring}}$ (ring region around the mask), $\Omega_{\text{bg}}$ (background region), and compute the optical flow statistics for each:
\begin{equation}
  c_t = Q_{q_c}(m_t | \Omega_{\text{core}}), \quad r_t = S(m_t | \Omega_{\text{ring}}), \quad b_t = S(m_t | \Omega_{\text{bg}}), \quad \beta_t = \max(r_t, b_t)
\end{equation}
where $Q$ denotes the quantile function and $S$ denotes the mean or quantile (controlled by \texttt{base stat}). Residual motion score:
\begin{equation}
  f_t = \max(0, c_t - \beta_t) + \alpha \cdot c_t
\end{equation}
with $\alpha = 0.3$. Adding the strain term to capture non-rigid and boundary changes:
\begin{equation}
  e_t = Q_{0.9}(|\nabla m_t|_2 \mid \Omega_{\text{core}}), \quad s_t = f_t + \lambda e_t
\end{equation}
with $\lambda = 0.2$. Optionally, size normalization for small targets: $f_t \leftarrow f_t / d_t \cdot d_{\text{ref}}$.

The final entity score jointly considers motion intensity and persistence:
\begin{equation}
  S_{90} = Q_{0.9}(\{s_t\}), \quad \tau = \max(\tau_{\text{floor}}, \rho \cdot S_{90}), \quad p = \frac{1}{T}\sum_t \mathbf{1}[s_t \ge \tau]
\end{equation}
\begin{equation}
  S_{\text{obj}} = S_{90} \cdot p^\gamma
\end{equation}
with $\gamma = 0.5$. Optional fusion with bbox geometric motion:
\begin{equation}
  g_t = 0.7(1 - \text{IoU}_t) + 0.3 \frac{|\mathbf{c}_{t+1} - \mathbf{c}_t|_2}{d_t}
\end{equation}
\begin{equation}
  S_{\text{final}} = S_{\text{obj}} + \eta \cdot Q_{0.9}(\{g_t\}) \cdot \max(1, S_{90})
\end{equation}
with $\eta = 0.2$.

\noindent\textbf{Design rationale:}
(1)~$c_t - \beta_t$: suppresses common-mode motion from background/camera;
(2)~$+\alpha c_t$: prevents over-suppression from eliminating genuine small-target motion;
(3)~$+\lambda e_t$: supplements non-rigid and boundary change information;
(4)~$S_{90} \times p^\gamma$: requires motion to be both ``strong'' and ``persistent,'' suppressing sporadic spikes;
(5)~geometric term: provides stable supplementary evidence when optical flow is unreliable.

\subsection{Dataset Cases}
\label{sec:appendix_dataset_cases}

\begin{figure}[t]
    \centering
    \includegraphics[width=\textwidth]{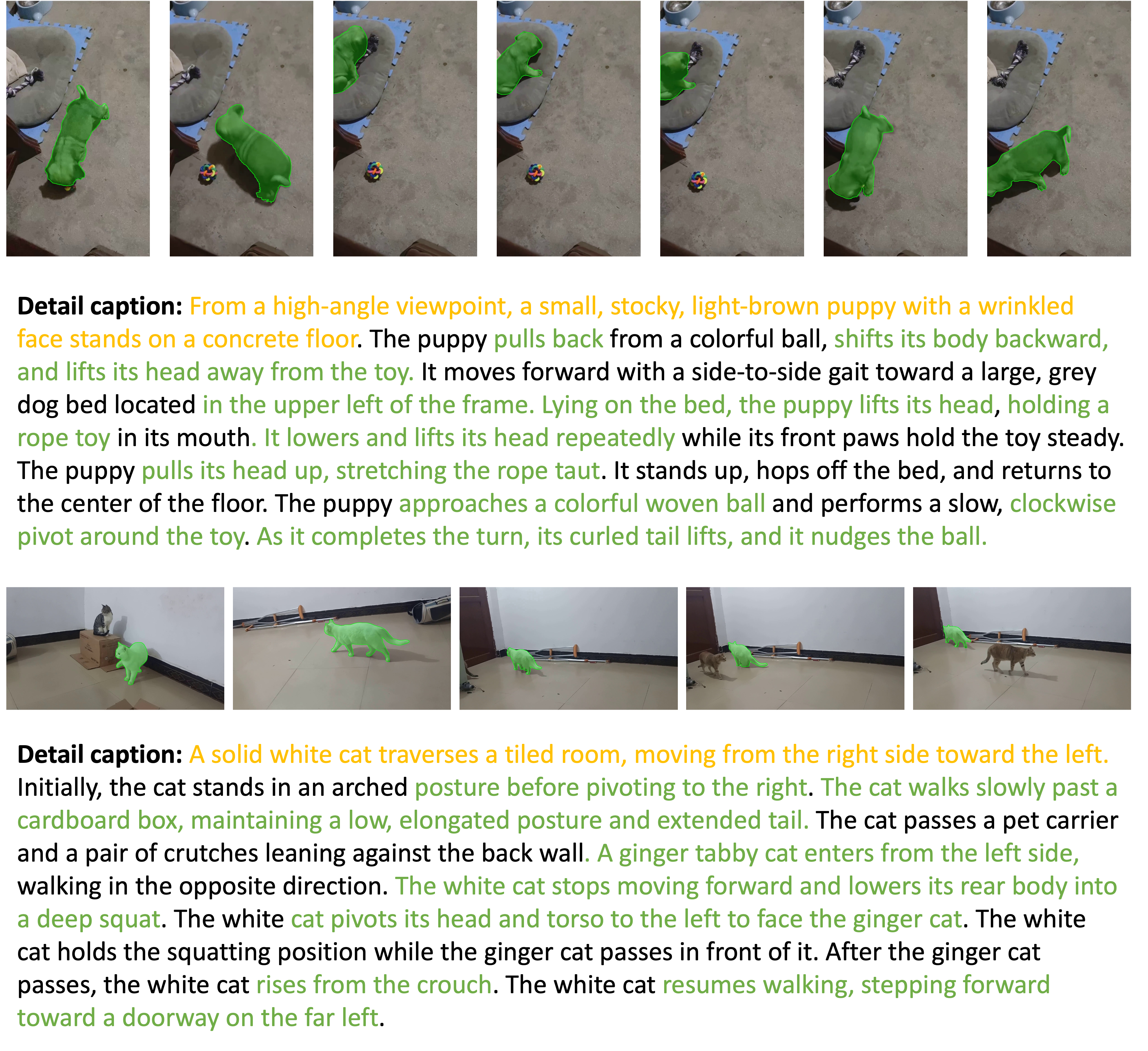}
    \caption{Qualitative examples from our dataset. The detailed captions convey comprehensive temporal sequences and rich descriptions of dynamic motion.}
    \label{fig:dataset_cases}
\end{figure}

To demonstrate the quality of the annotations, we present qualitative examples from our dataset in Figure~\ref{fig:dataset_cases}. As illustrated, our detailed captions possess sufficient temporal and dynamic descriptions. They accurately capture the fine-grained motion sequences, posture changes, and spatial transitions of the referent objects. By explicitly delineating semantic actions in a coherent, sequential manner, these captions provide reliable, comprehensive text supervision for complex spatiotemporal understanding.

\section{Supplementary Experiments}
\label{sec:appendix_exp}

\subsection{Data Source Effectiveness Ablation}
\label{sec:appendix_data_source}

To understand the impact of different data sources on fine-grained motion understanding, we conduct a controlled ablation by training Qwen3-VL-4B on 10K samples from three distinct data sources: (1)~MotionAtlas-Data (ours), (2)~VideoRefer~\cite{yuan2025videorefer} detail caption data, and (3)~FAVOR-Bench~\cite{tu2025favorbench} data. All other training settings are kept identical.

\begin{table}[t]
\centering\small
\caption{Data source effectiveness ablation. We compare 10K samples from three data sources using Qwen3-VL-4B as the base model.}
\label{tab:data_source_ablation}
\vspace{-10pt}
\setlength{\tabcolsep}{5pt}
\resizebox{\textwidth}{!}{
\begin{tabular}{l cccc}
\toprule
\textbf{Training Data (10K)} & \textbf{MotionAtlas} (Ours) & \textbf{MotionBench}~\cite{hong2025motionbench} & \textbf{DREAM-1K} F1~\cite{wang2024tarsierrecipestrainingevaluating} & \textbf{TOMATO}~\cite{shangguan2025tomato} \\
\midrule
Qwen3-VL-4B (baseline)~\cite{bai2025qwen3vltechnicalreport} & 19.20 & 55.90 & 35.90 & 27.36 \\
\midrule
~~+ 10K MotionAtlas-Data & \textbf{21.85} \up{} & 57.40 \up{} & 36.50 \up{} & \textbf{28.12} \up{} \\
~~+ 10K VideoRefer-125K~\cite{yuan2025videorefer} & 18.42 \down{} & 55.13 \down{} & 35.26 \down{} & 26.81 \down{} \\
~~+ 10K FAVOR-Bench Data~\cite{tu2025favorbench} & 20.96 \up{} & \textbf{58.10} \up{} & \textbf{36.80} \up{} & 27.89 \up{} \\
\bottomrule
\end{tabular}}
\end{table}

As shown in Table~\ref{tab:data_source_ablation}, MotionAtlas-Data consistently improves over the baseline across all four benchmarks, with the best results on MotionAtlas-Bench and TOMATO. In contrast, VideoRefer detail caption data 125K underperforms the baseline on all benchmarks. We conjecture that this is because VideoRefer is primarily appearance-oriented: although it contains detailed descriptions, its supervision is biased more toward object attributes and visual appearance than fine-grained motion dynamics, making it less effective for motion-centric training.

Compared with FAVOR-Bench, a high-quality human-annotated data source, MotionAtlas-Data achieves competitive overall performance at the same scale, and performs better on benchmarks requiring finer-grained motion reasoning. This shows that our automated pipeline can produce high-quality supervision comparable to human annotation, while offering much stronger scalability.

\subsection{Judge Robustness}
\label{sec:appendix_judge}

Our benchmark scores captions using MCQ-based fact verification rather than pairwise caption ranking. A pilot study found that fact verification reduces style and model-preference bias, but it still relies on a capable judge that faithfully follows the caption evidence. We observed that Gemini 2.5 Pro and Gemini 3 Pro follow caption evidence more faithfully than Qwen3-VL-235B, so we adopt Gemini 2.5 Pro as the default judge.

To verify that our conclusions do not depend on this particular choice, we re-score all models with alternative judges and measure the Spearman $\rho$ of the resulting model ranking against the default judge. Judge swaps preserve the ranking well: Gemini 3 Pro ($\rho=0.98$), GPT-5.2 ($\rho=0.96$), and Qwen3-VL-235B ($\rho=0.89$). This confirms that the relative ordering of models on MotionAtlas-Bench is robust to the choice of judge.

\subsection{Frame Sampling Ablation}
\label{sec:appendix_fps}

To investigate the impact of temporal density on fine-grained motion understanding, we evaluate multiple models on our MotionAtlas-Bench using 8, 16, and 32 uniformly sampled frames. The results for both Single-Frame Grounding and Full-Sequence Grounding settings are reported in Table~\ref{tab:frame_ablation}.

\begin{table}[t]
\centering\small
\caption{Frame sampling ablation on MotionAtlas-Bench (Accuracy). Performance at 8, 16, and 32 frames across different model scales.}
\label{tab:frame_ablation}
\vspace{-10pt}
\setlength{\tabcolsep}{3pt}
\resizebox{\textwidth}{!}{
\begin{tabular}{l ccccc}
\toprule
\multirow{2}{*}{\textbf{Frames}} & \multicolumn{5}{c}{\textbf{Single-Frame Grounding}} \\
\cmidrule(lr){2-6}
& Gemini 3 Pro & Qwen3-VL-235B & Qwen3-VL-32B & Qwen3-VL-8B & Ours (Qwen3-VL-8B) \\
\midrule
8 frames  & 34.82 & 29.40 & 28.50 & 23.15 & 30.12 \\
16 frames & 36.42 & 30.54 & 29.68 & 24.31 & 31.64 \\
32 frames & 37.15 & 31.45 & 28.85 & 23.40 & 32.50 \\
\midrule
\multirow{2}{*}{\textbf{Frames}} & \multicolumn{5}{c}{\textbf{Full-Sequence Grounding}} \\
\cmidrule(lr){2-6}
& Gemini 3 Pro & Qwen3-VL-235B & Qwen3-VL-32B & Qwen3-VL-8B & Ours (Qwen3-VL-8B) \\
\midrule
8 frames & 35.21 & 32.10 & 34.20 & 25.50 & 32.80 \\
16 frames & 36.47 & 33.74 & 35.70 & 26.72 & 34.05 \\
32 frames & 37.20 & 34.30 & 34.60 & 25.45 & 35.15 \\
\bottomrule
\end{tabular}}
\end{table}

As shown in Table~\ref{tab:frame_ablation}, all models exhibit higher accuracy when increasing the frame count from 8 to 16, confirming that a richer temporal context is fundamental for capturing complex motion dynamics. However, as the frame count increases to 32, a divergence emerges: while massive models like Gemini 3 Pro and Qwen3-VL-235B-A22B continue to improve, the standard Qwen3-VL-32B and Qwen3-VL-8B models experience noticeable performance degradation. This drop at higher frame counts is largely attributed to a decline in their object recognition and visual grounding capabilities. That is, the influx of denser spatial-temporal information overwhelms their ability to consistently track and refer to specific target regions. 

Importantly, after fine-tuning on our dataset (Ours-Qwen3-VL-8B), the model regains the ability to benefit from denser frame sampling, with accuracy increasing continuously up to 32 frames. This demonstrates that training on our region-aware data explicitly strengthens long-sequence object grounding, mitigating the contextual distraction issues observed in baseline models under high frame densities.

\paragraph{Native Frame Budgets.}
The 16-frame setting is our controlled comparison for fairness across models. When evaluated under their native or recommended frame budgets, proprietary and large open-source models improve further on MotionAtlas-Bench: Gemini 3 Pro (native) gains \up{4.8}, GPT-5.2 (50f) gains \up{6.3}, and Qwen3-VL-235B (native) gains \up{3.2}. The relative ordering is preserved, indicating that our controlled 16-frame protocol does not disadvantage any particular model family.

\paragraph{Temporal Evidence vs.\ Frame Count.}
Our benchmark targets the ability to exploit temporal evidence in highly dynamic motion, not merely longer video duration. Using the latest strong model, we conduct a matched 16f-vs-32f study with Gemini 3.1 Pro: MotionAtlas-Bench improves by \up{1.2}; for context, TOMATO~\cite{shangguan2025tomato} improves by \up{1.6}, while DREAM-1K~\cite{wang2024tarsierrecipestrainingevaluating} slightly decreases by \down{0.8}. This shows that MotionAtlas-Bench rewards genuine temporal evidence rather than penalizing or saturating with more frames.

\paragraph{Why Region-Grounded Training Transfers.}
This frame-density behavior also explains the cross-benchmark transfer reported in the main text (Table~\ref{tab:transfer_ablation}). We find that MotionAtlas-Data contains fine-grained motion cues that are answerable at 32 frames but \textit{not} at 16 frames, \textit{i.e.}, the supervision explicitly requires reading temporally dense evidence. Training on such data therefore teaches models to exploit the temporal density that base models underuse, which is consistent with the larger transfer gains observed when region-detail captioning and visual region cues are added.

\section{Broader Impacts and Ethics Statement}
\label{sec:ethics}

This work introduces MotionAtlas, a benchmark and dataset designed to advance fine-grained referential motion understanding. All source videos utilized in our dataset construction are sampled from existing, publicly available datasets (e.g., DREAM-1K, MeViS). 
We have ensured that our data curation process strictly focuses on motion dynamics and physical interactions, avoiding the intentional collection or annotation of personally identifiable information (PII) or sensitive biometric data.

While our automated pipeline leverages large language models (e.g., Gemini 3 Pro) for scalable annotation, we acknowledge that VLMs can occasionally hallucinate or reflect inherent biases in their pretraining data. 
To mitigate this risk, we implemented a rigorous, multi-stage self-bootstrap refinement and human-verification protocol to ensure high factual accuracy. 

In terms of broader societal impact, advancing precise motion understanding has significant positive applications in embodied AI, robotics, and autonomous driving, where fine-grained motion perception is crucial for safety. 
However, we also acknowledge that highly capable region-aware tracking and understanding technologies could, in theory, be misused for unauthorized surveillance. 
We encourage the computer vision community to utilize this dataset responsibly and to continue developing safeguards against the potential misuse of detailed video analysis tools.

\end{document}